\pdfoutput=1

\documentclass[11pt]{article}

\usepackage[final]{emnlp2021}

\usepackage{times}
\usepackage{latexsym}

\usepackage[T2A,T1]{fontenc}

\usepackage[utf8]{inputenc}

\usepackage{microtype}

\usepackage{url}

\usepackage{amsmath,amsfonts,bm}

\def\eqref#1{equation~\ref{#1}}

\def\1{\bm{1}}

\DeclareMathAlphabet{\mathsfit}{\encodingdefault}{\sfdefault}{m}{sl}
\SetMathAlphabet{\mathsfit}{bold}{\encodingdefault}{\sfdefault}{bx}{n}

\usepackage{hyperref}
\usepackage{graphicx}
\usepackage{caption}
\usepackage{subcaption}
\usepackage{physics}
\usepackage{enumitem}
\usepackage{booktabs}
\usepackage{mathtools}
\usepackage{footmisc}
\usepackage{makecell}
\usepackage{xcolor}
\usepackage{xspace}
\newcounter{srCounter}
\newif\ifsrvar
\srvartrue
\ifsrvar
\newcommand{\seb}[1]{{\small \color{red} \refstepcounter{srCounter}\textsf{[SR]$_{\arabic{srCounter}}$:{#1}}}}
\else
\newcommand{\seb}[1]{}
\fi

\newcounter{fpCounter}
\newif\iffpvar
\fpvartrue
\iffpvar
\newcommand{\fabio}[1]{{\small \color{blue} \refstepcounter{fpCounter}\textsf{[FP]$_{\arabic{fpCounter}}$:{#1}}}}
\else
\newcommand{\fabio}[1]{}
\fi

\newcounter{trCounter}
\newif\iftrvar
\trvartrue
\iftrvar
\newcommand{\tim}[1]{{\small \color{purple} \refstepcounter{trCounter}\textsf{[TR]$_{\arabic{trCounter}}$:{#1}}}}
\else
\newcommand{\tim}[1]{}
\fi

\newcounter{apCounter}
\newif\ifapvar
\apvartrue
\ifapvar
\newcommand{\piktus}[1]{{\small \color{orange} \refstepcounter{apCounter}\textsf{[AP]$_{\arabic{apCounter}}$:{#1}}}}
\else
\newcommand{\piktus}[1]{}
\fi

\newcounter{plCounter}
\newif\ifplvar
\plvartrue
\ifplvar
\newcommand{\patrick}[1]{{\small \color{magenta} \refstepcounter{plCounter}\textsf{[PL]$_{\arabic{plCounter}}$:{#1}}}}
\else
\newcommand{\patrick}[1]{}
\fi

\newcounter{afCounter}
\newif\ifafvar
\afvartrue
\ifafvar
\newcommand{\angela}[1]{{\small \color{olive} \refstepcounter{afCounter}\textsf{[AF]$_{\arabic{afCounter}}$:{#1}}}}
\else
\newcommand{\angela}[1]{}
\fi

\newcounter{kpCounter}
\newif\ifkpvar
\kpvartrue
\ifkpvar
\newcommand{\kpopat}[1]{{\small \color{olive} \refstepcounter{kpCounter}\textsf{[KP]$_{\arabic{kpCounter}}$:{#1}}}}
\else
\newcommand{\kpopat}[1]{}
\fi

\newcounter{ndcCounter}
\newif\ifndcvar
\ndcvartrue
\ifndcvar

\else
\newcommand{\ndc}[1]{}
\fi

\newcounter{ncCounter}
\newif\ifncvar
\ncvartrue
\ifncvar
\newcommand{\ncan}[1]{{\small \color{green} \refstepcounter{ncCounter}\textsf{[NC]$_{\arabic{ncCounter}}$:{#1}}}}
\else
\newcommand{\nc}[1]{}
\fi

\newcounter{lwCounter}
\newif\ifncvar
\ncvartrue
\ifncvar
\newcommand{\ledell}[1]{{\small \color{brown} \refstepcounter{lwCounter}\textsf{[LW]$_{\arabic{lwCounter}}$:{#1}}}}
\else
\newcommand{\ledell}[1]{}
\fi

\newcommand{\citepossessive}[1]{\citeauthor{#1}'s \citeyearpar{#1}}

\newcommand{\eg}{\textit{e.g.}}
\newcommand{\ie}{\textit{i.e.}}
\def\mgenre{m\textsc{GENRE}\@\xspace}
\def\genre{\textsc{GENRE}\@\xspace}

\title{Multilingual Autoregressive Entity Linking}
\author{
Nicola De Cao\textsuperscript{1,2}, \
Ledell Wu\textsuperscript{1}, \
Kashyap Popat\textsuperscript{1}, \\
{\bf Mikel Artetxe\textsuperscript{1}, \
Naman Goyal\textsuperscript{1}, \ 
Mikhail Plekhanov\textsuperscript{1}, \
Luke Zettlemoyer\textsuperscript{1,3}}, \\
{\bf Nicola Cancedda\textsuperscript{1}, \
Sebastian Riedel\textsuperscript{1,4}, \
Fabio Petroni\textsuperscript{1}} \\
\textsuperscript{1}Facebook AI \
\textsuperscript{2}University of Amsterdam \\
\textsuperscript{3}University of Washington \
\textsuperscript{4}University College London \\
\href{mailto:nicola.decao@gmail.com}{\texttt{{\color{black} nicola.decao@gmail.com}}} \\
\texttt{\{ledell, kpopat, artetxe, naman, movb} \\
\texttt{lsz, ncan, sriedel, fabiopetroni\}@fb.com}
}

\begin{document}
\maketitle
\begin{abstract}

We present \mgenre, a sequence-to-sequence system for the Multilingual Entity Linking (MEL) problem---the task of resolving language-specific mentions to a multilingual Knowledge Base~(KB).
For a mention in a given language, \mgenre predicts the name of the target entity left-to-right, token-by-token in an autoregressive fashion. 
The autoregressive formulation allows us to effectively cross-encode mention string and entity names to capture more interactions than the standard dot product between mention and entity vectors. It also enables fast search within a large KB even for mentions that do not appear in mention tables and with no need for large-scale vector indices.   
While prior MEL works use a single representation for each entity, we match against entity names of as many languages as possible, which allows exploiting language connections between source input and target name.
Moreover, in a zero-shot setting on languages with no training data at all, \mgenre treats the target language as a latent variable that is marginalized at prediction time. This leads to over 50\% improvements in average accuracy.
We show the efficacy of our approach through extensive evaluation including experiments on three popular MEL benchmarks where \mgenre establishes new state-of-the-art results. Code and pre-trained models at \url{https://github.com/facebookresearch/GENRE}.

\end{abstract}

\section{Introduction} \label{sec:intro}
Entity Linking~\citep[EL,][]{hoffart-etal-2011-robust,dredze-etal-2010-entity,bunescu-pasca-2006-using,cucerzan-2007-large} is an important task in NLP, with plenty of applications in multiple domains, spanning Question Answering~\citep{de-cao-etal-2019-question,nie-etal-2019-revealing,Asai2020Learning}, Dialogue~\citep{bordes2016learning,wen-etal-2017-network,williams-etal-2017-hybrid,chen-etal-2017-robust,curry2018alana}, Biomedical systems~\citep{f1852247075c44dc8a462b92b17a55f4,zheng2015entity}, to name just a few. It consists of grounding entity mentions in unstructured texts to KB descriptors (\eg, Wikipedia articles).

The multilingual version of the EL problem has been for a long time tight to a purely cross-lingual formulation~\citep[XEL,][]{mcnamee-etal-2011-cross,ji2015overview}, where mentions expressed in one language are linked to a KB expressed in another (typically English).
Recently,~\citet{botha-etal-2020-entity} made a step towards a more inherently multilingual formulation by defining a language-agnostic KB, obtained by grouping language-specific descriptors per entity. Such formulation has the power of considering entities that do not have an English descriptor (\eg, a Wikipedia article in English) but have one in some other languages.

\begin{figure*}[t!]
    \centering
    \includegraphics[width=\textwidth]{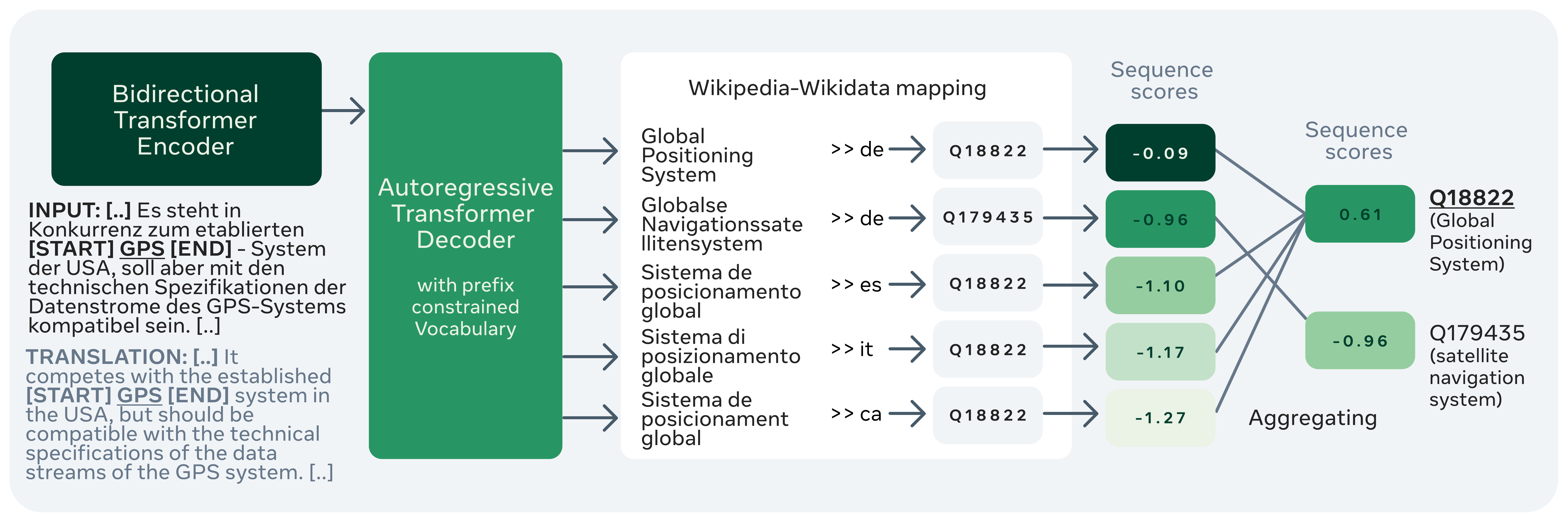}
    \caption{mGENRE: we use an autoregressive decoder to generate language IDs as well as entity names (\ie, Wikipedia titles). The combination of language ID and a entity name uniquely identify a Wikidata ID (with a N-to-1 mapping). We use Beam Search for efficient inference and we marginalize the probability scores for different languages to score entities. This example is a real output from our system.}
    \label{fig:mGENRE}
\end{figure*}

A common design choice to most current solutions, regardless of the specific formulation, is to provide a unified entity representation, either by collating multilingual descriptors in a single vector or by defining a canonical language. For the common bi-encoder approach ~\citep{wu-etal-2020-scalable, botha-etal-2020-entity}, this might be optimal. However, in the recently proposed \genre model ~\citep{decao2021autoregressive}, an autoregressive formulation to the EL problem leading to stronger performance and considerably smaller memory footprints than bi-encoder approaches on monolingual benchmarks, the representations to match against are entity names (\ie, strings) and it's unclear how to extend those beyond a monolingual setting.

In this context, we find that maintaining as much language information as possible, hence providing multiple representations per entity, helps due to the connections between source language and entity names in different languages. We additionally find that using all available languages as targets and aggregating over the possible choices is an effective way to deal with a zero-shot setting where no training data is available for the source language.

Concretely, in this paper, we present \mgenre, the first multilingual EL system that exploits a sequence-to-sequence architecture to generate entity names in more than 100 languages left to right, token-by-token in an autoregressive fashion and conditioned on the context (see Figure~\ref{fig:mGENRE} for an outline of our system).
While prior works use a single representation for each entity, we maintain entity names for as many languages as possible, which allows exploiting language connections between source input and target name.
To summarize, this work makes the following contributions:

\begin{itemize}[topsep=2pt,itemsep=-4pt]
    \item Extend the catalog of entity names by considering all languages for each entry in the KB. %
    Storing the multilingual names index is feasible and cheap (\ie, 2.2GB for $\sim$89M names).

    \item Design a novel objective function that marginalizes over all languages to perform a prediction. This approach is particularly effective in dealing with languages not seen during fine-tuning ($\sim$50\% improvements). 
    
    \item Establish new state-of-the-art performance for the Mewsli-9~\citep{botha-etal-2020-entity}, TR2016\textsuperscript{hard}~\citep{tsai-roth-2016-cross} and TAC-KBP2015~\citep{ji2015overview} MEL datasets. 
    
    \item Present extensive analysis of modeling choices, including the usage of candidates from a mention table, frequency-bucketed evaluation, and performance on an heldout set including low-resource languages.
    
    \item Publicly release our best model, pre-trained as multilingual denoising auto-encoder using the %
    BART objective \cite{lewis2019bart,liu2020multilingual} on large-scale monolingual corpora in 125 languages and fine-tuned to generate entity names given $\sim$730M in-context Wikipedia hyperlinks in 105 languages.

\end{itemize}
\section{Background}
We first introduce Multilingual Entity Linking in Section~\ref{sec:mel} highlighting its difference with monolingual and cross-lingual linking.
We address the MEL problem with a sequence-to-sequence model that generates textual entity identifiers (\ie, entity names). Our formulation generalizes the \genre model by~\citet{decao2021autoregressive} to a multilingual setting (\mgenre). Thus in Section~\ref{sec:genre} and~\ref{sec:genre_bs}, we discuss the \genre model and how it ranks entities with Beam Search respectively.

\subsection{Task Definition} \label{sec:mel}

\textit{Multilingual Entity Linking}~\citep[MEL,][]{botha-etal-2020-entity} is the task of linking a given entity mention $m$ in a given context $c$ of language $l \in \mathcal{L}_C$ to the corresponding entity $e \in \mathcal{E}$ in a multilingual Knowledge Base (KB).
See Figure~\ref{fig:mGENRE} for an example: there are textual inputs with entity mentions (in bold) and we ask the model to predict the corresponding entities in the KB.
A language-agnostic KB includes at least the name (but could include also descriptions, aliases, etc.) of each entity in one or more languages but there is no assumption about the relationship between these languages $\mathcal{L}_{KB}$ and languages of the context $\mathcal{L}_C$. This is a generalization of both monolingual Entity Linking EL and cross-lingual EL~\citep[XEL,][]{mcnamee-etal-2011-cross,ji2015overview}. The latter considers contexts in different languages while mapping mentions to entities in a monolingual KB (\eg, English Wikipedia).

Additionally, we assume that each $e \in \mathcal{E}$ has a unique textual identifier in at least a language. Concretely, in this work, we use Wikidata~\citep{vrandevcic2014wikidata} as our KB. Each Wikidata item lists a set of Wikipedia pages in multiple languages linked to it and in any given language each page has a unique name (\ie, its title).

\subsection{Autoregressive generation} \label{sec:genre}
\genre ranks each $e \in \mathcal{E}$ by computing a score with an autoregressive formulation: $\mathrm{score}_\theta(e|x) = p_\theta(y|x) = \prod_{i=1}^N p_\theta(y_i|y_{<i},x)$ where $y$ is the sequence of $N$ tokens in the identifier of $e$, $x$ the input (\ie, the context $c$ and mention $m$), and $\theta$ the parameters of the model. \genre is based on fine-tuned BART architecture~\citep{lewis2019bart} and it is trained using a standard seq2seq objective, \ie, maximizing the output sequence likelihood with teacher forcing~\citep{sutskever2011generating,sutskever2014sequence} and regularized with dropout~\citep{JMLR:v15:srivastava14a} and label smoothing~\citep{szegedy2016rethinking}.

\subsection{Ranking with Constrained Beam Search} \label{sec:genre_bs}
At test time, it is prohibitively expensive to compute a score for every element in $\mathcal{E}$ and then sort them. Thus, \genre exploits Beam Search~\citep[BS,][]{sutskever2014sequence}, an established approximate decoding strategy to navigate the search space efficiently. Instead of explicitly scoring all entities in $\mathcal{E}$, search for the top-$k$ entities in $\mathcal{E}$ using BS with $k$ beams. BS only considers one step ahead during decoding (\ie, it generates the next token conditioned on the previous ones). Thus, \genre employs a prefix tree (trie) to enable constrained beam search and then generate only valid entity identifiers.

\section{Model}

\label{sec:training}
To extend GENRE to a multilingual setting we need to define what are the unique identifiers of all entities in a language-agnostic fashion. This is not trivial since we rely on text representations that are by their nature grounded in some language. Concretely, for each entity $e$, we have a set of identifiers $\mathcal{I}_e$ that consists of pairs $\langle l, n_e^l \rangle$ where $l \in \mathcal{L}_{KB}$ indicates a language and $n_e^l$ the name of the entity $e$ in the language $l$. We extracted these identifiers from our KB---each Wikidata item has a set of Wikipedia pages in multiple languages linked to it, and in any given language, each page has a unique name. We identified 3 strategies to employ these identifiers:
\begin{enumerate}[label=\roman*),topsep=2pt,itemsep=-4pt]
    \item define a \textit{canonical} textual identifier for each entity such that there is a 1-to-1 mapping between the two (\ie, for each entity, select a specific language for its name---see Section~\ref{sec:canonical});
    \item define a N-to-1 mapping between textual identifier and entities concatenating a language ID (\eg, a special token or the ISO 639-1 code\footnote{\url{https://www.iso.org/standard/22109.html}}) followed by its name in that language---alternatively concatenating its name first and then a language ID (see Section~\ref{sec:plain_generation});
    \item treat the selection of an identifier in a particular language as a latent variable (\ie, we let the model learn a conditional distribution of languages given the input and we marginalize over those---see Section~\ref{sec:marginalization}).
\end{enumerate}
All of these strategies define a different way we compute the underlining likelihood of our model. In Figure~\ref{fig:mGENRE} we show an outline of \mgenre. The following subsections will present detailed discussions of the above 3 strategies.

\subsection{Canonical entity representation} \label{sec:canonical}
Selecting a single textual identifier for each entity corresponds to choosing its name among all the available languages of that entity.
We employ the same data-driven selection heuristic as in~\citet{botha-etal-2020-entity}: for each entity $e$ we sort all its names $n_e^l$ for each language $l$ according to the number of mentions of $e$ in documents of language $l$. Then we take the name $n_e^l$ in the language $l$ that has the most mentions of $e$. In case of a tie, we select the language that has the most number of mentions across all entities (\ie, the language for which we have more training data).
Having a single identifier for each entity corresponds to having a 1-to-1 mapping between strings and entities. Thus, $\mathrm{score}_\theta(e|x) = p_\theta(n_e|x)$ where with $n_e$ we indicate the \textit{canonical} name for $e$. We train to maximize the scores for all our training data. A downside of this strategy is that most of the time, the model 
cannot exploit the lexical overlap between the context and entity name since it has to translate it in the canonical one. 

\subsection{Multilingual entity representation} \label{sec:plain_generation}
To accommodate the canonical representation issues, we can predict entity names in any language.
Concatenating a language ID $l$ and an entity name $n_e^l$ in different orders induces two alternative factorizations. We train maximizing the scores for all our training data: $\mathrm{score}_\theta(e|x) =$
\begin{equation}
\begin{cases}
p_\theta(l|x) \cdot p_\theta(n_e^l | x, l) & \text{for `lang+name'} \\
p_\theta(n_e^l| x) \cdot p_\theta(l|n_e^l,x) & \text{for `name+lang'} \\
\end{cases}
\end{equation}
The former corresponds to first predicting a distribution over languages and then predicting a title \textit{conditioning} on the language $l$ where the latter corresponds to the opposite.
Predicting the language first conditions the generation to a smaller set earlier during beam search (\ie, all names in a specific language). However, it might exclude some targets from the search too early if the beam size is too small.
Predicting the language last does not condition the generation of names in a particular language but it \textit{asks} the model to disambiguate the language of the generated name whenever it is ambiguous (\ie, when the same name in different languages corresponds to possibly different entities). Only 1.65\% of the entity names need to be disambiguated with the language.
In practice, we observe no difference in performance between the two approaches.
Both strategies define an N-to-1 mapping between textual identifiers and entities and then at test time we just use a lookup table to select the correct KB item. This N-to-1 mapping is an advantage compared to using canonical names because the model can predict in any available language and therefore exploit synergies between source and target language as well as avoiding translation.

\subsection{Marginalization} \label{sec:marginalization}
Differently from the plain generation strategies described above, we can treat the textual identifiers as a latent variable and express $\mathrm{score}_\theta(e|x)$ as the probability of the entity name in all languages and marginalizing over them: $\mathrm{score}_\theta(e|x)=$
\begin{equation}
    p_\theta(e|x) = \sum_{\langle l, n_e^l \rangle \in \mathcal{I}_e} p_\theta(n_e^l,l|x) \;.
\end{equation}
Marginalization exposes the model to all representations in all languages of the same entity and it requires a minor modification of the training procedure. Unfortunately, because computing $\mathrm{score}_\theta(e|x)$ requires a sum over all languages, both training, and inference with marginalization are more expensive than with simple generation (scaling linearly with the number of languages).
However, at least during inference, we can still apply BS to only marginalize using the top-k generations.
For this reason, we test this training strategy only on few languages but we evaluate marginalization even when training with the other generation strategies described above.

\subsection{Candidate selection} \label{sec:candidates}
Modern EL systems that employ cross-encoding between context and entities usually do not score all entities in a KB as it is too computational expensive~\citep{wu-etal-2020-scalable}. Instead, they first apply candidate selection (with a less expensive method first or just a non-parametric mention table) to reduce the number of entities before scoring. In our formulation, there is no need to do that since \mgenre uses Beam Search to efficiently generate. However, using candidates might help, and therefore, we also experiment with that. Scoring all candidates might not be always possible (sometimes there are thousands of candidates for a mention) and especially when using an N-to-1 mapping between textual identifiers there will be names to rank in all languages available for each candidate. Thus, when we use candidates, it is to constrain BS steps further, rather than to rank all of them.

Concretely, candidate selection is made with an alias table. Using the training data, we build a mention table where we record all entities indexed by the names used to refer to them.
Additionally, we also use Wikipedia titles as additional mentions (useful for entities that never appear as links), redirects, Wikidata labels, and aliases.

\section{Experimental Setting}

We use Wikidata~\citep{vrandevcic2014wikidata} as our KB while exploiting the supervision signal from Wikipedia hyperlinks. For evaluation, we test our model on two established cross-lingual datasets, TR2016\textsuperscript{hard} and TAC-KBP2015~\citep{ji2015overview, tsai-roth-2016-cross}, as well as the recently proposed Mewsli-9 MEL dataset~\citep{botha-etal-2020-entity}.
Additionally, we propose a novel setting extracted from Wikinews\footnote{\url{https://www.wikinews.org}} where we train a model on a set of languages, and we test it on unseen ones.

\subsection{Knowledge Base: Wikidata}
We use Wikidata as the target KB to link to filtering the same heuristic as~\citet{botha-etal-2020-entity} (see Appedinx~\ref{app:experiments} for more details).
Eventually, our entity set $\mathcal{E}$ contains 20,277,987 items (as a reference, English Wikipedia has just $\sim$6M items). Using the corresponding Wikipedia titles as textual identifiers in all languages leads to a table of 53,849,351 entity names. We extended the identifiers including redirects which leads to a total of 89,270,463 entity names (see Table~\ref{tab:titles105} in Appendix~\ref{app:experiments} for more details).
Although large, the number of entity names is not a bottleneck as the generated prefix tree only occupies 2.2GB for storage. As a comparison the
\citepossessive{botha-etal-2020-entity} MEL systems need $\sim$10 times more storage.

\begin{figure*}[t]
    \centering
    \includegraphics[width=\textwidth]{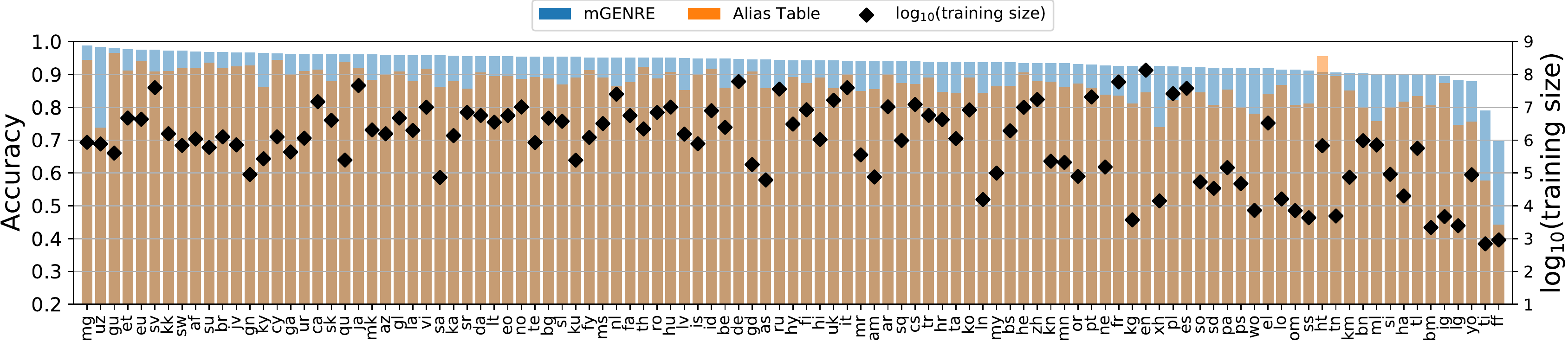}
    \caption{Accuracy of mGENRE on the 105 languages in our Wikipedia validation set. We also report the accuracy of the alias table and the log-training set sizes per each language (see Figure~\ref{fig:stats105_langs_large} and Table~\ref{tab:stats105} in Appendix~\ref{app:additonal} for a larger view and for precise values and language full names).}
    \label{fig:stats105_langs}
\end{figure*}

\subsection{Supervision: Wikipedia}
For all experiments, we do not train a model from scratch, but we fine-tune a multilingual language model trained on 125 languages (see Appendix~\ref{app:experiments} for more details on the pre-trained model).
We exploit Wikipedia hyperlinks as the source of supervision for MEL. We used Wikipedia in 105 languages out of the $>$300 available. These 105 are all the languages for which our model was pre-trained on that overlaps with the one available in Wikipedia (see full language list in Figure~\ref{fig:stats105_langs} and more details in Appendix~\ref{app:experiments}).
Eventually, we extracted a large-scale dataset of 734,826,537 datapoints.
For the plain generation strategy, we selected as the ground truth the name in the source language. When such entity name is not available we randomly select 5 alternative languages and we use all of them as datapoints.
To enable model selection, we randomly selected 1k examples from each language to keep as a validation set. 

\subsection{Datasets
} \label{sec:data_test}

\paragraph{Mewsli-9} \citep{botha-etal-2020-entity} contains 289,087 entity mentions appearing in 58,717 originally written news articles from
Wikinews, linked to WikiData. The corpus includes documents in 9 languages.\footnote{Arabic, English, Farsi, German, Japanese, Serbian, Spanish, Tamil, and Turkish.} Differently from the cross-lingual setting, this is a truly multilingual dataset since 11\% target entities in Mewsli-9 do not have an English Wikipedia page thus a XEL model would not link these.

\paragraph{TR2016\textsuperscript{hard}} \citep{tsai-roth-2016-cross} is a Wikipedia based cross-lingual dataset specifically constructed to contain difficult mention-entity pairs. Authors extracted Wikipedia hyperlinks for which the corresponding entity is not the most likely when using an alias table. Since we train on Wikipedia, to avoid an overlap with this test data, we removed all mentions from our training data that also appear in TR2016\textsuperscript{hard}. Note that this pruning strategy is more aggressive than~\citepossessive{tsai-roth-2016-cross} and~\citepossessive{botha-etal-2020-entity} strategies.
\citet{tsai-roth-2016-cross} assured to not have mention-entity pairs overlaps between training and test, but a mention (with a different entity) might appear in training.
\citet{botha-etal-2020-entity}\footnote{Information provided by private correspondence with the authors.} split at the page-level only, making sure to hold out all \citet{tsai-roth-2016-cross} test pages (and their corresponding pages in other languages), but they trained on any mention-entity pair that could be extracted from their remaining training page partition (\ie, they have overlap between training and text entity-mention pairs).
To compare with previous works~\citep{tsai-roth-2016-cross, upadhyay-etal-2018-joint, botha-etal-2020-entity} we only evaluate on German, Spanish, French and Italian (a total of 16,357 datapoints).

\paragraph{TAC-KBP2015} To evaluate our system on documents out of the Wikipedia domain, we experiment on the TAC-KBP2015 Tri-Lingual Entity Linking Track~\citep{ji2015overview}. To compare with previous works~\citep{tsai-roth-2016-cross, upadhyay-etal-2018-joint, sil2018neural, zhou-etal-2019-towards}, we use only Spanish and Chinese (\ie, we do not evaluate in English). Following previous work, we only evaluate \textit{in-KB} links~\citep{yamada-etal-2016-joint,ganea-hofmann-2017-deep}, i.e, we do not evaluate on mentions that link to entities out of the KB. Previous works considered Freebase~\citep{bollacker2008freebase} as KB, and thus we computed a mapping between Freebase ID and Wikidata ID. When we cannot solve the match, our system gets zero scores (\ie, it counts as a wrong prediction). TAC-KBP2015 contains 166 Chinese documents (84 news and 82 discussion forum articles) and 167 Spanish documents (84 news and 83 discussion forum articles) for a total of 12,853 mention-entity datapoints.

\paragraph{Wikinews-7} For the purpose of testing a model on languages unseen during training, we extract mention-entities pairs from Wikinews in 7 languages that are not in the Mewsli-9 language set.\footnote{Chinese, Czech, French, Italian, Polish, Portuguese, and Russian.} Table~\ref{tab:unseen_stats} in Appendix~\ref{app:experiments} reports statistics of this dataset. Wikinews-7 is created in the same way as Mewsli-9, but we used our own implementation to extract data from raw dumps\footnote{\citet{botha-etal-2020-entity} did not release code for extracting Mewsli-9 from a Wikinews dump.}.

\section{Results} \label{sec:results}

\begin{table*}[t]
\small
\centering
\begin{tabular}{r cccccc}
\toprule
& \multicolumn{2}{c}{\citet{botha-etal-2020-entity}} & \multicolumn{4}{c}{\bf Ours} \\
\cmidrule(lr){2-3} \cmidrule(lr){4-7}
{\bf Language} & Alias Table & Model F\textsuperscript{+} & mGENRE  & + cand.&  + marg. & + cand. + marg. \\
\midrule
ar        &                      89.0 &            92.0 &           94.7 &              94.8 &              95.3 &             \textbf{95.4} \\
de        &                      86.0 &   \textbf{92.0} &                       91.5 &                          91.8 &                          91.8 &             \textbf{92.0} \\
en        &                      79.0 &            87.0 &                       86.7 &              87.1 &                          87.0 &             \textbf{87.2} \\
es        &                      82.0 &            89.0 &           90.0 &     \textbf{90.1} &     \textbf{90.1} &             \textbf{90.1} \\
fa        &                      87.0 &            92.0 &  \textbf{94.6} &     \textbf{94.6} &              94.2 &                      94.4 \\
ja        &                      82.0 &            88.0 &           89.9 &              91.1 &              90.2 &             \textbf{91.4} \\
sr        &                      87.0 &            93.0 &           94.9 &              94.4 &     \textbf{95.0} &                      94.5 \\
ta        &                      79.0 &            88.0 &           92.9 &              93.3 &              93.1 &             \textbf{93.8} \\
tr        &                      80.0 &            88.0 &           90.7 &              91.4 &              90.9 &             \textbf{91.5} \\
\midrule
\textbf{micro-avg} &                      83.0 &            89.0 &           90.2 &              90.5 &              90.4 &             \textbf{90.6} \\
\textbf{macro-avg} &                      83.0 &            90.0 &           91.8 &              92.1 &              92.0 &             \textbf{92.3} \\
\bottomrule
\end{tabular}
\caption{Accuracy on Mewsli-9 dataset. We report results of \mgenre (trained with `title+lang') with and without top-k candidates from the table as well as with and without marginalization. 
}
\label{tab:mewsli9}
\end{table*}

\begin{table*}[t]
\small
\centering
\begin{tabular}{lccc|ccccc}
\toprule
& \multicolumn{3}{c}{\textbf{TAC-KBP2015}} & \multicolumn{5}{c}{\textbf{TR2016\textsuperscript{hard}}} \\
\cmidrule(lr){2-4} \cmidrule(lr){5-9}
\textbf{Method}&\textbf{es}&\textbf{zh} & \textbf{macro-avg} &    \textbf{de} &    \textbf{es} &    \textbf{fr} &    \textbf{it}& \textbf{macro-avg} \\
\midrule
\citet{tsai-roth-2016-cross} & 82.4 & 85.1 & 83.8 & 53.3 & 54.5 & 47.5 & 48.3 & 50.9\\
\citet{sil2018neural}* & 83.9 & 85.9 & 84.9 & - & - & - & - & -\\
\citet{upadhyay-etal-2018-joint} & 84.4 & 86.0 & 85.2 & 55.2 & 56.8 & 51.0 & 52.3 & 53.8\\
\citet{zhou-etal-2019-towards} & 82.9 & 85.5 & 84.2 & - & - & - & - & -\\
\citet{botha-etal-2020-entity} & - & - & - & \textbf{62.0} & 58.0 & 54.0 & 56.0 & 57.5\\
\midrule
\textbf{mGENRE} &  86.3 & 64.6 & 75.5 &  56.3 &  57.1 &  50.0 &  51.0 &   53.6 \\
\textbf{mGENRE} + marg.& \textbf{86.9} & 65.1 & 76.0 &  56.2 &  56.9 &  49.7 &  51.1 &   53.5 \\
\textbf{mGENRE} + cand.& 86.5 & 86.6 & 86.5 & 61.8 &  \textbf{61.0} &  \textbf{54.3} &  \textbf{56.9} &  \textbf{58.5} \\
\textbf{mGENRE} + cand. + marg. & 86.7 & \textbf{88.4} & \textbf{87.6} & 61.5 & 60.6 & \textbf{54.3} & 56.6 & 58.2 \\
\bottomrule
\end{tabular}
\caption{Accuracy on TAC-KBP2015 Entity Linking dataset (only datapoints linked to FreeBase) and TR2016\textsuperscript{hard} of \mgenre (trained with `title+lang') with and without top-k candidates from the table as well as with and without marginalization. *as reported by~\citet{upadhyay-etal-2018-joint}.}
\label{tab:tr2016_tackbp2015_results}
\end{table*}

The main results of this work are reported in Table~\ref{tab:mewsli9} for Mewsli-9, and in Table~\ref{tab:tr2016_tackbp2015_results} for TR2016\textsuperscript{hard}, and TAC-KBP2015 respectively. Our \mgenre (trained with `title+lang') outperforms all previous works in all those datasets. We show the accuracy of \mgenre on the 105 languages in our Wikipedia validation set against an alias table baseline in Figure~\ref{fig:stats105_langs}. In Table~\ref{tab:stats105} in Appendix~\ref{app:experiments} we report more details on the validation set results. 
In Table~\ref{tab:examples} we report some examples of correct and wrong predictions of our \mgenre on selected datapoints from Mewsli-9. 

\subsection{Performance evaluation}

\paragraph{Mewsli-9}
In Table~\ref{tab:mewsli9} we compare our \mgenre against the best model from~\citet{botha-etal-2020-entity} (Model~F\textsuperscript{+}) as well as with their alias table baseline. We report results from \mgenre with and without constraining the beam search to the top-k candidates from the table (see Section~\ref{sec:candidates}) as well as with and without marginalization (see Section~\ref{sec:marginalization}). All of these alternatives outperform Model~F\textsuperscript{+} on both micro and macro average accuracy across the 9 languages. Our base model (without candidates or marginalization) has a 10.9\% error reduction in micro average and 18.0\% error reduction for macro average over all languages. The base model has no restrictions on candidates so it is effectively classifying among all the $\sim$20M entities. The base model performs better than Model~F\textsuperscript{+} on each individual language except English and German. Note that these languages are the ones for which we have more training data ($\sim$134M and $\sim$60M datapoints each) but also the languages that have the most entities/pages ($\sim$6.1M and $\sim$2.4M). Therefore these are the hardest languages to link.\footnote{Even though there are many datapoints, disambiguating between more entities is harder. Moreover, low-resource languages should be harder as there are fewer training instances, but there are no such languages in the Mewsli-9 set.} When enabling candidate filtering to restrict the space for generation, we further improve error reduction to 13.6\% and 21.0\% for micro and macro average respectively. Marginalization reduces the error by the same amount as candidate filtering but combining search with candidates and marginalization leads to our best model: it improves error reduction to 14.5\% and  23.0\% on micro and macro average respectively. Our best model is also better than Model~F\textsuperscript{+} in English and on par with it in German.

\paragraph{TR2016\textsuperscript{hard} and TAC-KBP2015}
We compared our \mgenre against cross-lingual systems~\citep{tsai-roth-2016-cross,sil2018neural,upadhyay-etal-2018-joint,zhou-etal-2019-towards} and Model~F\textsuperscript{+} by~\citet{botha-etal-2020-entity} in Table~\ref{tab:tr2016_tackbp2015_results}. Differently from Meswli-9, the base \mgenre model does not outperform previous systems. Using marginalization brings minimal improvements. Instead, using candidates gives +11\% absolute accuracy on TAC-KBP2015 and +5\% on TR2016\textsuperscript{hard} effectively making \mgenre state-of-the-art in both datasets. The role of candidates is very evident on TAC-KBP2015 where there is not much of a difference for Spanish but a +22\% absolute accuracy for Chinese. TAC-KBP2015 comes with a training set and we used it to expand the candidate set. Additionally, we also included all simplified Chinese versions of the entity names because we used traditional Chinese in pre-training, and TAC-KBP2015 uses simplified Chinese. Many mentions in  TAC-KBP2015 were not observed in Wikipedia, so the performance gain mostly comes from this but including the simplified and alternative Chinese names also played an important role (+5\% comes from this alone).\footnote{We speculate that including different version (\eg, different dialects for Arabic) of entity names could improve performance in all languages. Since this is not in the scope of this paper, we will leave it for future work.}

\begin{table}[t]
\centering
\small
\begin{tabular}{lrrrr}
\toprule
& \multicolumn{2}{c}{\citet{botha-etal-2020-entity}} & \multicolumn{2}{c}{\bf mGENRE} \\
\cmidrule(lr){2-3} \cmidrule(lr){4-5}
\textbf{Bin} & Support & Acc. & Support & Acc.\\
\midrule
{[}0, 1)    &               3,198 &            8.0 &               1,244 &           22.1 \\
{[}1, 10)   &               6,564 &           58.0 &               5,777 &           47.3 \\
{[}10, 100) &              32,371 &           80.0 &              28,406 &           77.3 \\
{[}100, 1k) &              66,232 &           90.0 &              72,414 &           89.9 \\
{[}1k, 10k) &              78,519 &           93.0 &              84,790 &           93.2 \\
{[}10k, +)  &             102,203 &           94.0 &              96,456 &           96.3 \\
\midrule
\textbf{micro-avg}   &             289,087 &           89.0 &             289,087 &           90.6 \\
\textbf{macro-avg}   &                  - &           70.0 &                  - &           71.0 \\
\bottomrule
\end{tabular}

\caption{Results on the Mewsli-9 dataset, by entity frequency in training. The support is slightly different because training data differ (\ie, the set of languages from Wikipedia is different).}
\label{tab:mewsli_bins}
\end{table}

\subsection{Analysis}

\paragraph{By entity frequency}
Table~\ref{tab:mewsli_bins} shows a breakdown of Mewsli-9 accuracy by entity frequency in training for~\citepossessive{botha-etal-2020-entity} Model~F\textsuperscript{+} and \mgenre. Interestingly, our \mgenre has much higher accuracy (22\% vs 8\%) on unseen entities (\ie, the {[}0,1) bin). This is because our formulation can take advantage of copying names from the source, translating them or normalizing them. For example, an unseen person name should likely be linked to the entity with the same name. This is a powerful bias that gives the model advantage in these cases. On very rare entities (\ie, the {[}1,10) bin) our model performs worse than Model~F\textsuperscript{+}. Note that Model~F\textsuperscript{+} was trained specifically to tackle those cases (\eg, with hard negatives and frequency-based mini-batches) whereas our model was not. We argue that similar strategies can be applied to \mgenre to improve performance on rare entities, and we leave that to future work. The performance gap between Model~F\textsuperscript{+} and \mgenre on entities that appear more than 100 times in the training set is negligible.

\begin{figure}[t]
    \centering
    \includegraphics[width=0.48\textwidth]{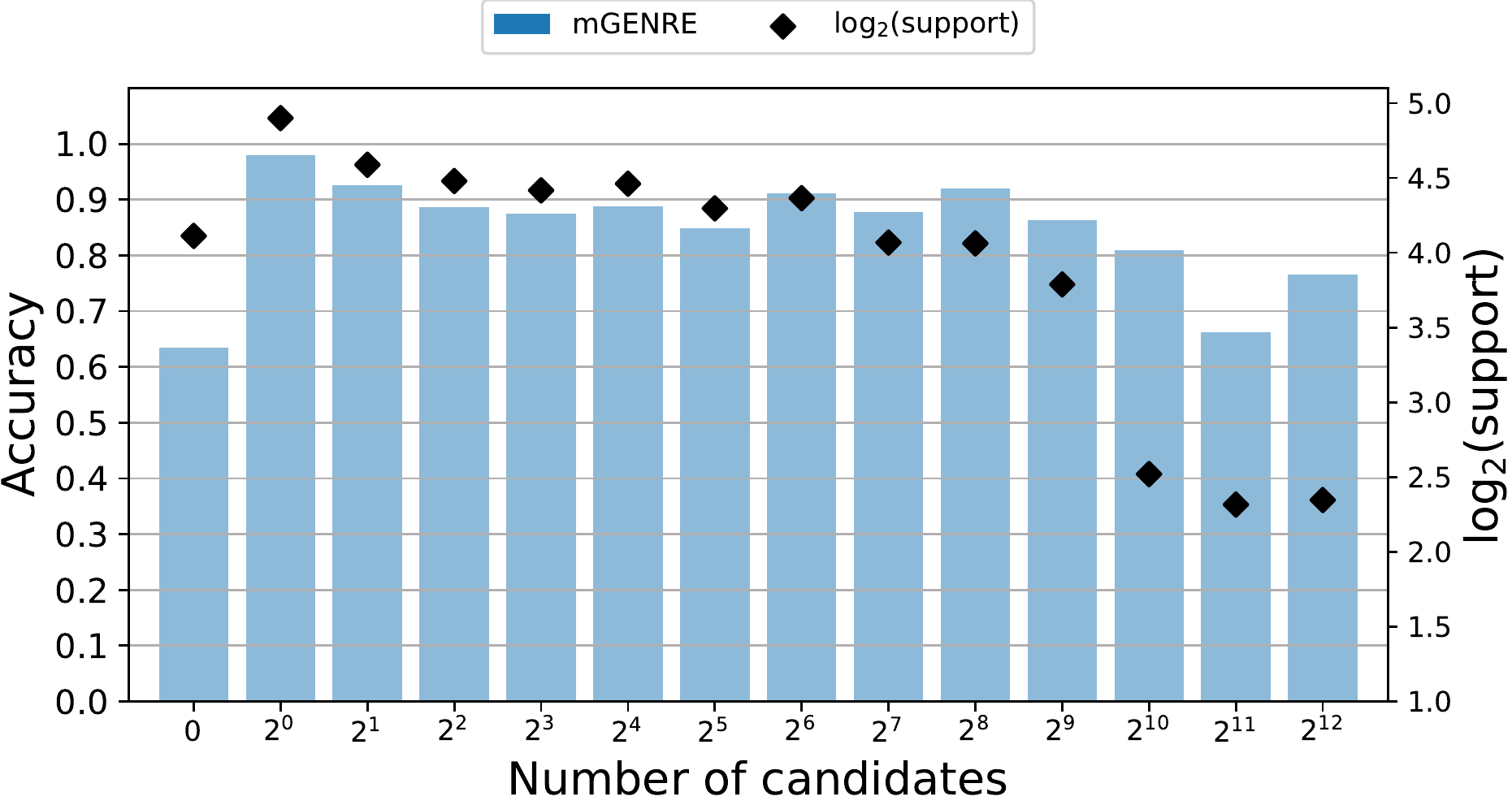}
    \caption{
    Results of \mgenre on Mewsli-9 by the number of retrieved candidates.
    }
    \label{fig:stats_mentions}
\end{figure}

\paragraph{By candidate frequency}
We additionally measure the accuracy on Mewsli-9 by the number of candidates retrieved from the alias table (details in Figure~\ref{fig:stats_mentions}). When there are no candidates ($\sim$12k datapoints that is $\sim$4\% Mewsli-9) an alias table would automatically fail, but \mgenre uses the entire KB as candidates and has 63.9\% accuracy. For datapoints with few candidates (\eg, less than 100), we could use \mgenre as a \textit{ranker} and score all of the options without relying on constrained beam search. However, this approach would be computationally infeasible when there are no candidates (\ie, we use all the KB as candidates) or too many candidates (\eg, thousands). Constrained BS allows us to efficiently explore the space of entity names, whatever the number of candidates.

\begin{table}[t]
\centering
\small
\begin{tabular}{r cccc} \toprule
{\bf Lang.} & {\bf Can.} & {\bf N+L}  & {\bf L+N} & {\bf L+N\textsuperscript{M}}  \\
\midrule
cs        &  36.3 &  30.2 &                      34.0 &          \textbf{69.7} \\
fr        &  62.9 &  57.0 &                      53.3 &          \textbf{73.4} \\
it        &  44.8 &  43.7 &                      42.9 &          \textbf{56.8} \\
pl        &  31.9 &  21.2 &                      25.6 &          \textbf{68.8} \\
pt        &  60.8 &  61.7 &                      59.5 &          \textbf{76.2} \\
ru        &  34.9 &  32.4 &                      35.1 &          \textbf{65.8} \\
zh        &  35.1 &  41.1 &                      44.0 &          \textbf{52.8} \\
\midrule
\textbf{micro-avg} &  41.6 &  38.3 &                      39.5 &          \textbf{65.9} \\
\textbf{macro-avg} &  43.8 &  41.0 &                      42.1 &          \textbf{66.2} \\
\bottomrule
\end{tabular}

\caption{\mgenre on the Wikinew-7 unseen languages. Models are trained only on the Mewsli-9 languages (1M datapoints per language). `Can.' is canonical, `N+L' is `name+language` and `L+N' is the opposite. \textsuperscript{M} indicates marginalization.}
\label{tab:unseen} 
\end{table}
\begin{figure}[t]
	\centering
	\begin{subfigure}[b]{0.48\textwidth}
		\centering
		\includegraphics[scale=0.6]{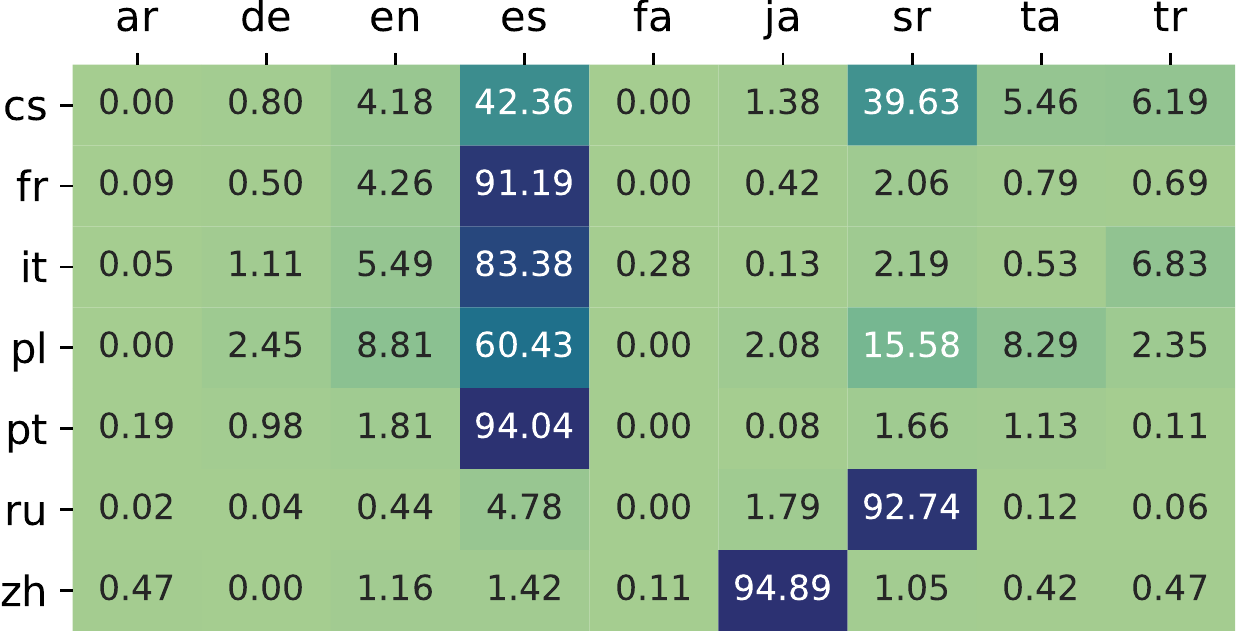}
		\caption{Lang+Name.}
		\label{fig:heatmaps_wikinews1}
	\end{subfigure}
    \par\vspace{6pt}
	\begin{subfigure}[b]{0.48\textwidth}
		\centering
		\includegraphics[scale=0.6]{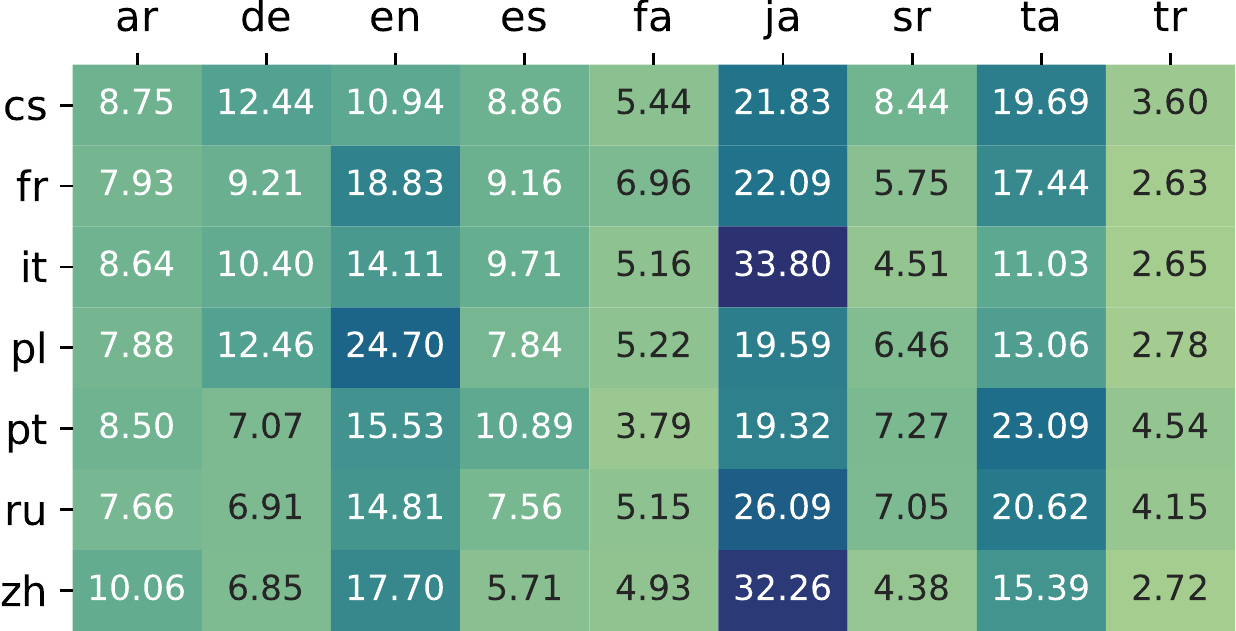}
		\caption{Lang+Name\textsuperscript{M}.}
		\label{fig:heatmaps_wikinews2}
	\end{subfigure}
	\caption{Distribution of languages on the top-1 prediction of two mGENRE models on Wikinews-7 (unseen languages). Y-axis indicates the source (unseen at training time) language where X-axis indicates the language (seen at training time) of the first prediction. %
	}
	\label{fig:heatmaps_wikinews}
\end{figure}

\paragraph{Unseen Languages}

We use our Wikinews-7 dataset to evaluate \mgenre capabilities to deal with languages not seen during training (\ie, the set of languages in train and test are disjoint).
This zero-shot setting implies that no mention table is available during inference; hence we do not consider candidates for test mentions.
We train our models on the nine Mewsli-9 languages and compare all strategies exposed in Section~\ref{sec:training}.
To make our ablation study feasible, 
we restrict the training data to the first 1 million hyperlinks from Wikipedia abstracts.
Results are reported in Table~\ref{tab:unseen}.

\begin{table*}[ht]
\centering
\resizebox*{!}{0.9\textwidth}{  

\fontsize{8.4}{10.1}\selectfont
\begin{tabular}{l p{.8\textwidth}} \toprule
  \textbf{Input} & 
  {[ES]}
  \ldots Michaëlle Jean, gobernadora general de Canadá, ha emitido el miércoles (13) un comunicado acerca del tiroteo ocurrido en el \textbf{[START] \underline{Dawson College} [END]} de Montreal. \ldots \\
  \textbf{Translation} &
  {[EN]}
  {\color{gray} \ldots Michaëlle Jean, Governor General of Canada, issued a statement on Wednesday (13) about the shooting that occurred at \textbf{[START] \underline{Dawson College} [END]} in Montreal. \ldots } \\
  \cmidrule(lr){2-2}
  \textbf{Prediction} &
  `\textit{Collège Dawson >> fr}': College in Montreal\textsuperscript{\href{https://www.wikidata.org/wiki/Q2983587}{Q2983587}} \\
  \cmidrule(lr){2-2}
  \textbf{Outcome} & \emph{\textbf{Correct}: mGENRE copies and normalizes the college name even if it is not does not have an identifier in the source language (i.e., it predicts in French but the source is Spanish).} \\
  \midrule
  \textbf{Input} &
  {[DE]}
  \ldots Etwa 47 Menschen sind bei den Protesten festgenommen worden, darunter Chas Booth, Mitglied der \textbf{[START] \underline{schottischen Grünen} [END]} mit Sitz im Stadtrat von Edinburgh. \ldots \\
  \textbf{Translation} &
  {[EN]}
  {\color{gray} \ldots Around 47 people were arrested during the protests, including Chas Booth, a member of the \textbf{[START] \underline{Scottish Greens} [END]} on the Edinburgh City Council. \ldots} \\
  \cmidrule(lr){2-2}
  \textbf{Prediction} &
  `\textit{Scottish Green Party >> de}': Scottish Green Party\textsuperscript{\href{https://www.wikidata.org/wiki/Q1256956}{Q1256956}} \\
  \cmidrule(lr){2-2}
  \textbf{Outcome} & \emph{\textbf{Correct}: even if the party is referred with its German alias, mGENRE predicts the identifier with its English name since the truth German Wikipedia page has the English name.} \\
  \midrule
  \midrule
  \textbf{Input} & 
  {[TR]}
  \ldots Kâinat Güzeli yine Venezuela\'dan \textbf{[START] \underline{2009 yılı Kâinat Güzellik Yarışması} [END]} 83 ülkenin temsilcisiyle Bahamalar\'da yapıldı. \ldots \\
  \textbf{Translation} &
  {[EN]}
  {\color{gray}\ldots Miss Universe is again from Venezuela \textbf{[START] \underline{The Universe Beauty Contest of 2009} [END]} was held in Bahamas with the representatives of 83 countries. \ldots } \\
  \cmidrule(lr){2-2}
  \textbf{Prediction} &
  `\textit{2009 Eurovision Çocuk Şarkı Yarışması >> tr}': Junior Eurovision Song Contest 2009\textsuperscript{\href{https://www.wikidata.org/wiki/Q205038}{Q205038}} \\
  \cmidrule(lr){2-2}
  \textbf{Annotation} & Miss Universe 2009\textsuperscript{\href{https://www.wikidata.org/wiki/Q756701}{Q756701}} \\
  \cmidrule(lr){2-2}
  \textbf{Outcome} & \emph{\textbf{Wrong}: the model is conditioned early during beam search to start with `2009'. Thus, it does not effectively search sequences where the year is at the end missing the ground truth answer.} \\
  \midrule
  \textbf{Input} &
  {[SR]}
  \ldots\;\textbf{[START]}~{\fontencoding{T2A}\selectfont \textbf{\underline{Марко Стојановић}}} \textbf{[END]}\ {\fontencoding{T2A}\selectfont, председник Светске организације пантомимичара, каже:}\ldots\\
  \textbf{Translation} & 
  {[EN]}
  {\color{gray}\ldots \textbf{[START] \underline{Marko Stojanović} [END]}, President of the World Mime Organization, says: \ldots } \\
  \cmidrule(lr){2-2}
  \textbf{Prediction} &
  {\fontencoding{T2A}\selectfont `Марко Стојановић >> sr'} : Marko Stojanović (lawyer)\textsuperscript{\href{https://www.wikidata.org/wiki/Q12754975}{Q12754975}} \\
  \cmidrule(lr){2-2}
  \textbf{Annotation} & Marko Stojanović (actor)\textsuperscript{\href{https://www.wikidata.org/wiki/Q16099367}{Q16099367}} \\
  \cmidrule(lr){2-2}
  \textbf{Outcome} & \emph{\textbf{Wrong}: 
  from the context \textit{lawyer} is potentially more appropriate than \textit{actor}. This is the risk of not considering the full entity description (that might say that Marko is an actor and also the President of the World Mime Organization). Even if on average copying is an effective strategy, it does not always succeed (using the alias table on this example leads to a correct prediction).
  } \\
  \bottomrule
\end{tabular}
}
\caption{Examples of correct and wrong predictions of our \mgenre model on selected samples from Mewsli-9. With both correct and wrong predictions we highlight some specific behaviour of our model.}
\label{tab:examples}
\end{table*}

Using our novel marginalization strategy that aggregates (both at training and inference time) over all seen languages to perform the linking brings an improvement of over 50\% with respect to considering a single language. To deeper investigate the behaviour of the model in this setting, we compute the probability mass distribution over languages seen at training time for the first prediction (reported in Figure~\ref{fig:heatmaps_wikinews}). When marginalization is enabled (Figure~\ref{fig:heatmaps_wikinews2}) the distribution is more spread across languages since the model is trained to use all of them. Hence the model can exploit connections between an unseen language and all seen languages for the linking process, which drastically increases the accuracy.

\section{Related Work}
The most related works to ours are~\citet{decao2021autoregressive}, that proposed to use an autoregressive language model for monolingual EL, and~\citet{botha-etal-2020-entity} that proposes to extend the cross-lingual EL task to multilingual EL with a language-agnostic KB.
We provide an outline of the GENRE model proposed by~\citet{decao2021autoregressive} in Section~\ref{sec:genre} and~\ref{sec:genre_bs}. GENRE was applied not only to EL but also for joint mention detection and entity linking (still with an autoregressive formulation) as well as to page-level document retrieval across multiple Knowledge Intensive Language Tasks~\citep[KILT;][]{petroni2020kilt} \ie, fact-checking, open-domain question answering, slot filling, and dialog.
\citepossessive{botha-etal-2020-entity} Model F\textsuperscript{+} is a \textit{bi-encoder} model: it is based on two BERT-based~\citep{devlin-etal-2019-bert} encoders that outputs vector representations for contet and entities. Similar to~\citet{wu-etal-2020-scalable} they rank entities with a dot-product between these representations. Model F\textsuperscript{+} uses the description of entities as input to the entity encoder and title, document and mention (separated with special tokens) as inputs to the context encoder.
Bi-encoders solutions may be memory inefficient since they require to keep in memory big matrices of embeddings, although memory-efficient dense retrieval has recently received attention ~\citep{izacard2020memory,min2021neurips,lewis2021paq}.

Another widely explored line of work is Cross-Language Entity Linking~\citep[XEL;][]{mcnamee2011cross, cheng-roth-2013-relational}. XEL considers contexts in different languages while mapping mentions to entities in a monolingual KB (\eg, English Wikipedia).
\citet{tsai2016cross} used alignments between languages to train multilingual entity embeddings. They used candidate selection and then they re-rank them with an SVM using these embeddings as well as a set of features (based on the multilingual title, mention, and context tokens). \citet{sil2018neural} explored the use of more sophisticated neural models for XEL as well as~\citet{upadhyay-etal-2018-joint} who jointly modeled type information to boost performance. \citet{zhou-etal-2019-towards} propose improvements to both entity candidate generation and disambiguation to make better use of the limited data in low-resource scenarios.
Note that in this work we focus on \textit{multilingual} EL, not cross-lingual. XEL is limiting to a monolingual KB (usually English), where MEL is more general since it can link to entities that might not be necessary represented in the target monolingual KB but in any of the available languages.

\section{Conclusion}
In this work, we propose an autoregressive formulation to the multilingual entity linking problem. For a mention in a given language, our solution generates entity names left-to-right and token-by-token.
The resulting system, \mgenre, maintains entity names in as many languages as possible to exploit language connections and interactions between source mention context and target entity name. 
The constrained beam search decoding strategy enables fast search within a large set of entity names (\eg, the whole KB in multiple languages) with no need for large-scale dense indices.
We additionally design a novel objective function that marginalizes over all available languages to perform a prediction. We empirically show that this strategy is really effective in dealing with languages for which no training data is available (\ie, 50\%  improvements for languages never seen during training). Overall, our experiments show that \mgenre achieves new state-of-the-art performance on three popular multilingual entity linking datasets.

\section*{Acknowledgments}
Authors thank
Patrick Lewis,
Aleksandra Piktus,
for helpful discussions and technical support.

\bibliography{anthology,bibliography}
\bibliographystyle{acl_natbib}

\clearpage
\appendix
\section{Experimental Details} \label{app:experiments}

\subsection{Pre-training}
\begin{figure}[t]
    \centering
    \includegraphics[scale=0.35]{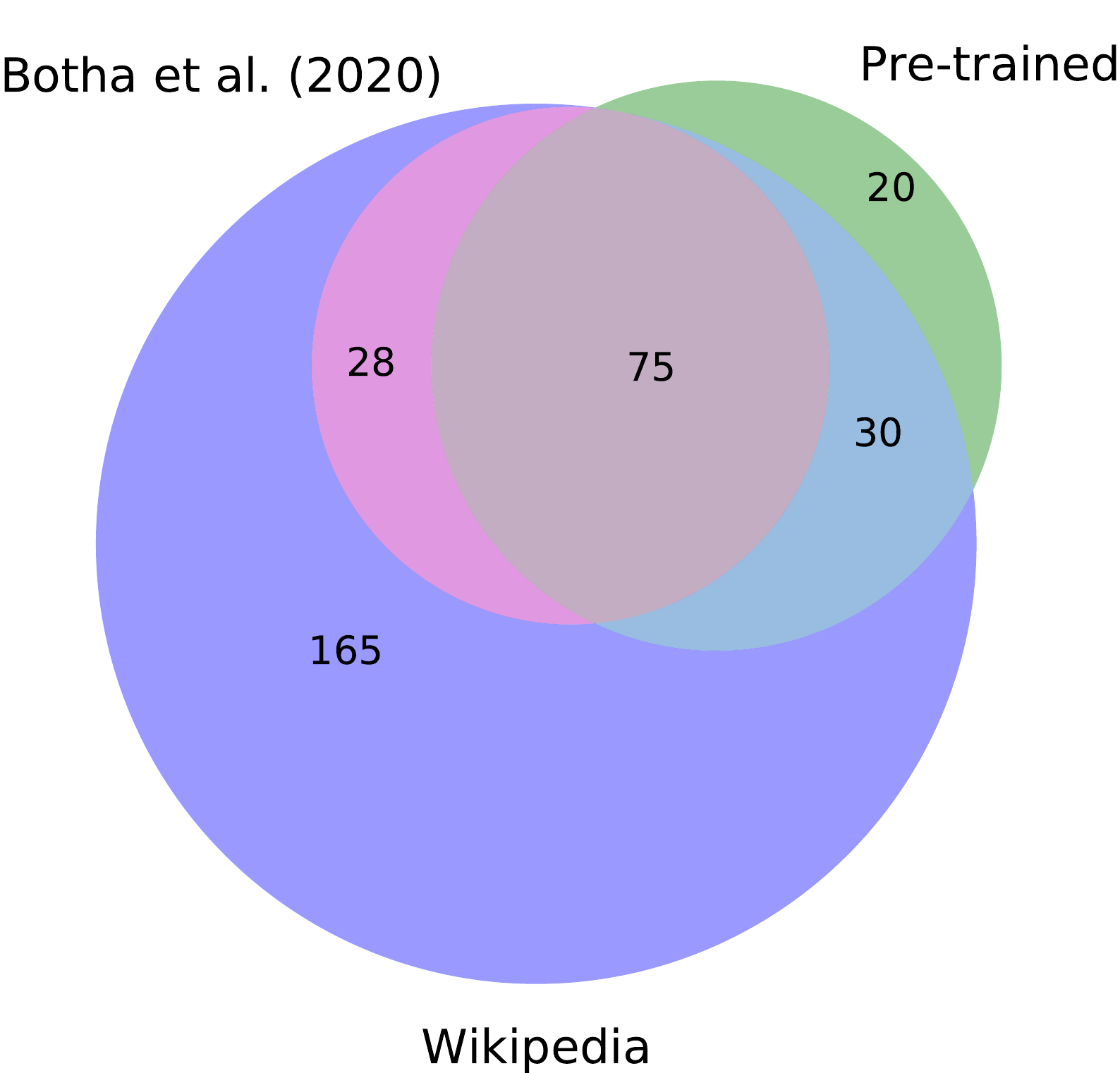}
    \caption{Venn diagram on the overlap of languages used during multilingual language modeling (pre-training), the languages available on Wikipedia (as of 2019-10-01), and the languages used by~\citet{botha-etal-2020-entity}. After pre-training on 125 languages, we fine-tune on the 105 that overlap with the one available in Wikipedia.}
    \label{fig:venn_langs}
\end{figure}

We used a pre-trained mBART~\citep{lewis2019bart,liu2020multilingual} model on 125 languages---see Figure~\ref{fig:venn_langs} for a visual overview of the overlap with these languages, Wikipedia and the languages used by~\citet{botha-etal-2020-entity}. mBART has 24 layers of hidden size is 1,024 and it has a total of 406M parameters.
We pre-trained on an extended version of the \texttt{cc100}~\citep{conneau-etal-2020-unsupervised,wenzek-etal-2020-ccnet} corpora available here\footnote{\url{http://data.statmt.org/cc-100}} where we increased the number of common crawl snapshots for low resource languages from 12 to 60.
The dataset has $\sim$5TB of text. We pre-trained for 500k steps with max 1,024 tokens per GPU on a variable batch size ($\sim$3000). Figure~\ref{fig:venn_langs} shows a Venn diagram on the overlap of languages used during pre-training and fine-tuning. 

\subsection{Data for supervision}

\paragraph{Wikidata}
Wikidata contains tens of millions of items but most of them are scholarly articles or they correspond to help and template pages in Wikipedia (\ie, not entities we want to retain) \footnote{\url{https://www.wikidata.org/wiki/Wikidata:Statistics}}.
Following~\citep{botha-etal-2020-entity}, we only keep Wikidata items that have an associated Wikipedia page in at least one language, independent of the languages we actually model. Moreover, we filter out items that are a subclass (P279) or instance of
(P31) some Wikimedia organizational entities (\eg, help and template pages---see Table~\ref{tab:wikidata_filter}).

\paragraph{Wikipedia}
We aligned each Wikipedia hyperlink to its respective Wikidata item using a custom script. Note that each Wikipedia page maps to a Wikidata item. For the alignment we use i) direct reference when the hyperlink point directly to a Wikipedia page, ii) a re-directions table if the hyperlink points to an alias page, and iii) a Wikidata search among labels and aliases of items if the previous two alignment strategies failed. The previous two alignment strategies might fail when i) authors made a mistake linking on a non-existing page, ii) authors linked to a non-existing page on purpose hoping it will be created in the future, or iii) the original title of a page changed over time and no redirection was added to accommodate old hyperlinks. This procedure successfully aligns 91\% of the hyperlinks. We only keep unambiguous alignments since, when using Wikidata search (\ie, the third alignment strategy), the mapping could be ambiguous (\eg, multiple items may share the same labels and aliases).

In Table~\ref{tab:titles105} we report some statistics of the training data extracted from Wikipedia.
We use a standard Wikipedia extractor \texttt{wikiextractor}\footnote{\url{https://github.com/attardi/wikiextractor}} by~\citet{Wikiextractor2015} and a redirect extractor\footnote{\url{https://code.google.com/archive/p/wikipedia-redirect}}.
We use both Wikipedia and Wikidata dumps from 2019-10-01.

\begin{table}[t]
\small
\centering
\begin{tabular}{l l} \toprule
{\bf Wikidata ID} & {\bf Label} \\
\midrule
Q4167836  &  category \\
Q24046192 &  category stub \\
Q20010800 &  user category \\
Q11266439 &  template \\
Q11753321 &  navigational template \\
Q19842659 &  user template \\
Q21528878 &  redirect page \\
Q17362920 &  duplicated page \\
Q14204246 &  project page \\
Q21025364 &  project page \\
Q17442446 &  internal item \\
Q26267864 &  KML file \\
Q4663903  &  portal \\
Q15184295 &  module \\
\bottomrule
\end{tabular}
\caption{Wikidata identifiers used for filtering out items from~\citet{botha-etal-2020-entity}.
\label{tab:wikidata_filter}}
\end{table}

\subsection{Data for test}
We use Wikinews (from 2019-10-01) to construct our \textit{unseen} Wikinews-7 dataset. In Table~\ref{tab:unseen_stats} we report some statistic of our new dataset.

\begin{table}[t]
\small
\centering
\begin{tabular}{lrrrr}
\toprule
& & & \multicolumn{2}{c}{\bf Entities} \\
\cmidrule(lr){4-5}
\textbf{Lang.} &  \textbf{Docs} &  \textbf{Mentions} &  Distinct &  $\not\in$ EnWiki \\
\midrule
ru &  1,625 &     20,698 &       8,832 &              1,838 \\
it &   907 &      8,931 &       4,857 &               911 \\
pl &  1,162 &      5,957 &       3,727 &               547 \\
fr &   978 &      7,000 &       4,093 &               349 \\
cs &   454 &      2,902 &       1,974 &               200 \\
pt &   666 &      2,653 &       1,313 &               113 \\
zh &   395 &      2,057 &       1,274 &               110 \\
\midrule
\textbf{Total} &   6,187 &      50,198 &       26,070 &               4,068 \\
\bottomrule
\end{tabular}
\caption{Corpus statistics for the Wikinews unseen languages we use as an evaluation set.}
\label{tab:unseen_stats}
\end{table}

\subsection{Training}
We implemented, trained, and evaluate our model using the \texttt{fariseq} library~\citep{ott2019fairseq}. We trained mGENRE using Adam~\citep{kingma2014adam} with a learning rate $10^{-4}$, $\beta_1=0.9$, $\beta_2=0.98$, and with a linear warm-up for 5,000 steps followed by liner decay for maximum 2M steps. The objective is sequence-to-sequence categorical cross-entropy loss with 0.1 of label smoothing and 0.01 of weight decay. We used dropout probability of 0.1 and attention dropout of 0.1. We used max 3,072 tokens per GPU and variable batch size ($\sim$12,500). Training was done on 384 GPUs (Tesla V100 with 32GB of memory) and it completed in $\sim$72h for a total of $\sim$27,648 GPU hours or $\sim$1,152 GPU days. Since TAC-KBP2015 contains noisy text (\eg, XML/HTML tags), we further fine-tune mGENRE for 2k steps on its training set when testing on it.

\subsection{Inference}
At test time, we use Constrained Beam Search with 10 beams, length penalty of 1, and maximum decoding steps of 32. We restrict the input sequence to be at most 128 tokens cutting the left, right, or both parts of the context around a mention. When employing marginalization, we normalize the log-probabilities by sequence length using $\log p(y|x) / L^\alpha$, where $\alpha=0.5$ was tuned on the development set.

\section{Additional results} \label{app:additonal}

\subsection{Analysis}
\paragraph{By mention frequency}
We show a breakdown of the accuracy of \mgenre on Mewsli-9 by mention frequency in Table~\ref{tab:mewsli_bins_mentions}. The accuracy of unseen mentions is 66.7\% and increases up to 93.6\% for mentions seen more than 10k times. For extremely common mentions (\ie, seen more than 1M times) the accuracy drops to 73.2\%.  
These mentions correspond to entities that are harder to disambiguate (\eg, `United States' appears 3.2M times but can be linked to the country as well as any sports team where the context refers to sports).

\paragraph{Unseen Languages}
Even though marginalization and canonical representation are the top-two systems in the unseen languages setting, they are not on seen languages. In Table~\ref{tab:ablation_mewsli} we report the results of all these strategies also on the seen languages (Mewsli-9 test set). Complementary to Figure~\ref{fig:heatmaps_wikinews} we also report the probability mass distribution over languages seen for Mewsli-9.

\begin{table}[t]
\small
\centering
\begin{tabular}{lrr}
\toprule
\textbf{Bin} & Support & Acc. \\
\midrule
{[}0, 1)      & 14,741 & 66.7 \\
{[}1, 10)     & 15,279 & 88.1 \\
{[}10, 100)   & 43,169 & 92.0 \\
{[}100, 1k)   & 75,927 & 91.7 \\
{[}1k, 10k)   & 80,329 & 91.5 \\
{[}10k, 100k) & 47,944 & 93.6 \\
{[}100k, 1M)  & 11,460 & 93.0 \\
{[}1M, 10M)   & 238 & 73.2 \\
\bottomrule
\end{tabular}
\caption{\mgenre results on Mewsli-9 dataset by mention
frequency in training.}
\label{tab:mewsli_bins_mentions}
\end{table}

\begin{table}[t]
\small
\centering
\begin{tabular}{r cccc} \toprule
{\bf Lang.} & {\bf Can.} & {\bf N+L} & {\bf L+N} & {\bf L+N\textsuperscript{M}}  \\
\midrule
ar&90.5&92.8&\textbf{92.9}&89.2\\
de&\textbf{84.6}&86.4&86.4&85.3\\
en&77.6&\textbf{79.3}&79.2&76.5\\
es&83.4&\textbf{85.5}&85.2&83.4\\
fa&91.6&90.7&\textbf{91.8}&88.2\\
ja&81.3&82.3&\textbf{82.8}&81.3\\
sr&91.5&92.7&\textbf{92.9}&92.5\\
ta&\textbf{92.8}&91.8&91.9&91.3\\
tr&\textbf{88.0}&87.7&87.3&86.0\\
\midrule
micro-avg&83.20&84.77&\textbf{84.80}&83.05\\
macro-avg&86.82&87.68&\textbf{87.82}&85.97\\
\midrule
\midrule
& \multicolumn{4}{c}{\bf + candidates} \\
\midrule
ar&94.4&94.5&\textbf{94.7}&93.0\\
de&89.4&\textbf{89.8}&\textbf{89.8}&89.3\\
en&83.6&83.8&\textbf{83.9}&82.4\\
es&87.7&88.2&\textbf{88.3}&87.3\\
fa&\textbf{93.6}&93.3&\textbf{93.6}&93.3\\
ja&87.9&88.0&\textbf{88.4}&87.9\\
sr&93.1&93.4&\textbf{93.5}&93.2\\
ta&\textbf{93.0}&92.2&92.5&92.5\\
tr&\textbf{91.1}&90.4&89.9&89.1\\
\midrule
micro-avg&87.95&88.22&\textbf{88.32}&87.43\\
macro-avg&90.42&90.41&\textbf{90.51}&89.78\\
\bottomrule
\end{tabular}
\caption{\mgenre on the Mewsli-9. Models are trained only on the Mewsli-9 languages (1M datapoints per language). `Can.' is canonical, `N+L' is `name+language` and `L+N' is the opposite. \textsuperscript{M} indicates marginalization.}
\label{tab:ablation_mewsli}
\end{table}

\begin{figure}[t]
	\centering
	\begin{subfigure}[b]{0.48\textwidth}
		\centering
		\includegraphics[scale=0.6]{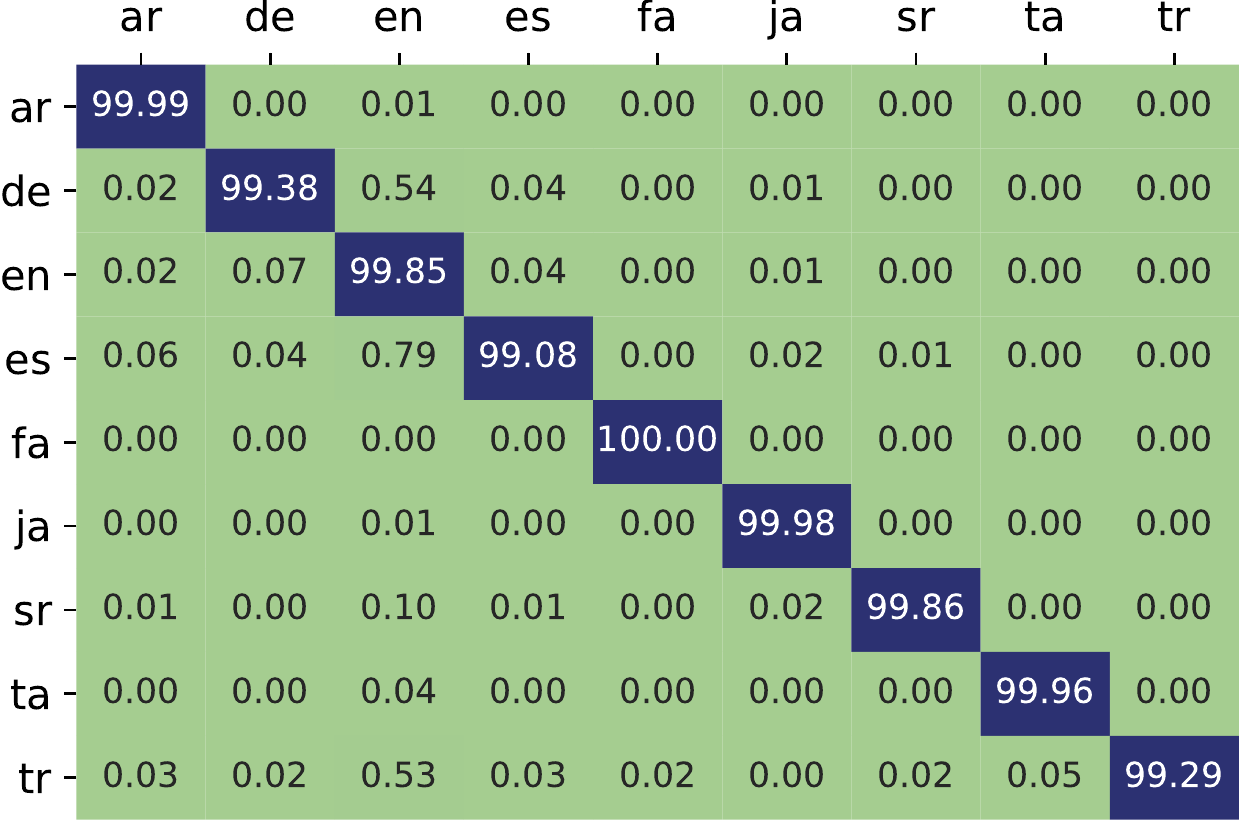}
		\caption{Lang+Name.}
		\label{fig:heatmaps_mewsli1}
	\end{subfigure}
	\par\vspace{6pt}
	\begin{subfigure}[b]{0.48\textwidth}
		\centering
		\includegraphics[scale=0.6]{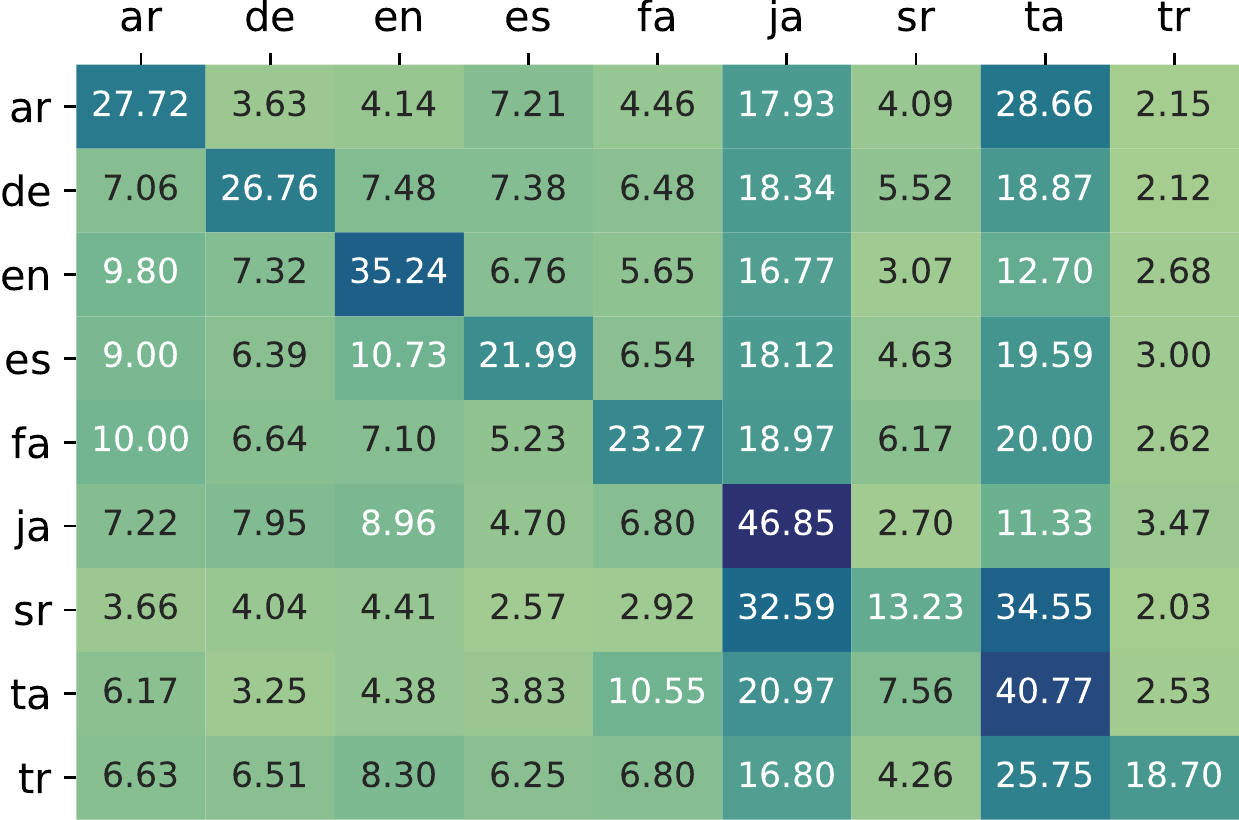}
		\caption{Lang+Name\textsuperscript{M}.}
		\label{fig:heatmaps_mewsli2}
	\end{subfigure}
	\caption{Distribution of languages on the top-1 prediction of two mGENRE models on Mewsli-9. Y-axis indicates the source language where X-axis indicates the language of the top-1 prediction. The models trained on those languages.}
	\label{fig:heatmaps_mewsli}
\end{figure}

\clearpage
\begin{table*}[t]
\centering
\resizebox*{!}{0.729\textheight}{  

\fontsize{8.4}{10.1}\selectfont
\begin{tabular}[t]{lrrr}
\toprule
\textbf{Language} & \textbf{Pages}  & \textbf{Names} & \textbf{Hyperlinks} \\
\midrule

Afrikaans (af)       & 85,456 & 110,705  & 1,089,581 \\
Albanian (sq)        & 86,234 & 112,112  & 978,394 \\
Amharic (am)         & 15,280 & 19,905  & 75,575 \\
Arabic (ar)          & 971,861 & 1,883,080  & 10,308,074 \\
Armenian (hy)        & 260,395 & 582,941  & 3,082,000 \\
Assamese (as)        & 6,119 & 19,041  & 61,209 \\
Azerbaijani (az)     & 152,033 & 189,793  & 1,562,968 \\
Bambara (bm)         & 747 & 916  & 2,191 \\
Basque (eu)          & 337,916 & 430,456  & 4,305,648 \\
Belarusian (be)      & 181,030 & 415,152  & 2,459,794 \\
Bengali (bn)         & 76,121 & 257,730  & 960,484 \\
Bosnian (bs)         & 82,164 & 184,148  & 1,916,515 \\
Breton (br)          & 67,388 & 88,284  & 1,255,295 \\
Bulgarian (bg)       & 257,962 & 376,934  & 4,655,641 \\
Burmese (my)         & 48,683 & 55,700  & 98,992 \\
Catalan (ca)         & 630,340 & 1,024,519  & 14,790,419 \\
Chinese (zh)         & 1,085,180 & 1,951,612  & 17,262,417 \\
Croatian (hr)        & 193,705 & 250,008  & 4,223,179 \\
Czech (cs)           & 439,249 & 719,643  & 12,173,376 \\
Danish (da)          & 255,957 & 405,745  & 5,621,483 \\
Dutch (nl)           & 1,986,801 & 2,714,649  & 25,002,389 \\
English (en)         & 6,071,492 & 14,751,661  & 134,477,329 \\
Esperanto (eo)       & 270,871 & 447,159  & 5,570,306 \\
Estonian (et)        & 201,505 & 342,215  & 4,700,888 \\
Finnish (fi)         & 470,896 & 737,165  & 8,390,037 \\
French (fr)          & 2,160,840 & 3,718,185  & 59,006,932 \\
Frysk (fy)           & 42,893 & 72,490  & 1,206,432 \\
Fulah (ff)           & 306 & 421  & 912 \\
Gaelic, (gd)         & 15,126 & 23,631  & 180,186 \\
Galician (gl)        & 159,849 & 229,561  & 4,709,070 \\
Ganda (lg)           & 2,376 & 2,668  & 2,476 \\
Georgian (ka)        & 135,040 & 138,267  & 1,369,094 \\
German (de)          & 2,356,465 & 3,877,850  & 60,638,345 \\
Greek (el)           & 170,541 & 251,692  & 3,310,875 \\
Guarani (gn)         & 3,755 & 5,589  & 89,593 \\
Gujarati (gu)        & 29,091 & 32,526  & 402,483 \\
Haitian (ht)         & 59,350 & 63,279  & 677,064 \\
Hausa (ha)           & 4,143 & 5,025  & 19,929 \\
Hebrew (he)          & 253,861 & 444,127  & 9,947,354 \\
Hindi (hi)           & 138,378 & 192,652  & 1,040,288 \\
Hungarian (hu)       & 459,261 & 663,995  & 10,138,904 \\
Icelandic (is)       & 48,563 & 75,963  & 772,213 \\
Igbo (ig)            & 1,521 & 3,000  & 4,702 \\
Indonesian (id)      & 516,196 & 1,015,784  & 7,882,254 \\
Irish (ga)           & 51,824 & 61,336  & 435,135 \\
Italian (it)         & 1,571,189 & 2,450,009  & 39,382,886 \\
Japanese (ja)        & 1,173,978 & 1,877,660  & 45,957,053 \\
Javanese (jv)        & 57,422 & 75,792  & 718,589 \\
Kannada (kn)         & 25,986 & 33,880  & 227,731 \\
Kazakh (kk)          & 229,165 & 271,260  & 1,564,344 \\
Khmer (km)           & 9,838 & 12,349  & 73,950 \\
Kongo (kg)           & 1,247 & 1,440  & 3,733 \\
Korean (ko)          & 475,605 & 1,061,961  & 8,309,492 \\
\bottomrule
\end{tabular}
}
~
\resizebox*{!}{0.75\textheight}{  

\fontsize{8.4}{10.1}\selectfont
\begin{tabular}[t]{lrrr}
\toprule
\textbf{Language} & \textbf{Pages}  & \textbf{Names} & \textbf{Hyperlinks} \\
\midrule
Kurdish (ku)         & 26,963 & 42,134  & 244,779 \\
Kyrgyz (ky)          & 80,985 & 89,486  & 271,335 \\
Lao (lo)             & 4,414 & 5,761  & 16,173 \\
Latin (la)           & 132,410 & 186,829  & 1,986,307 \\
Latvian (lv)         & 99,062 & 226,570  & 1,522,814 \\
Lingala (ln)         & 3,262 & 4,134  & 15,518 \\
Lithuanian (lt)      & 197,215 & 282,077  & 3,512,764 \\
Macedonian (mk)      & 103,960 & 152,384  & 2,035,348 \\
Malagasy (mg)        & 92,500 & 142,156  & 857,000 \\
Malay (ms)           & 331,403 & 388,110  & 3,190,700 \\
Malayalam (ml)       & 67,475 & 152,809  & 712,869 \\
Marathi (mr)         & 55,601 & 100,904  & 355,536 \\
Mongolian (mn)       & 21,772 & 28,455  & 208,847 \\
Nepali (ne)          & 34,107 & 39,904  & 151,958 \\
Norwegian (no)       & 521,665 & 816,772  & 10,234,086 \\
Oriya (or)           & 15,532 & 30,431  & 79,261 \\
Oromo (om)           & 1,063 & 1,317  & 7,153 \\
Panjabi (pa)         & 33,934 & 46,720  & 145,204 \\
Pashto (ps)          & 11,773 & 16,878  & 46,987 \\
Persian (fa)         & 716,604 & 2,139,255  & 5,567,774 \\
Polish (pl)          & 1,370,672 & 1,812,412  & 25,817,929 \\
Portuguese (pt)      & 1,053,673 & 1,858,821  & 20,625,904 \\
Quechua (qu)         & 21,670 & 41,230  & 247,508 \\
Romanian (ro)        & 403,517 & 979,524  & 6,974,837 \\
Russian (ru)         & 1,585,051 & 3,592,042  & 35,783,391 \\
Sanskrit (sa)        & 11,960 & 22,472  & 73,380 \\
Serbian (sr)         & 625,871 & 3,248,789  & 7,012,202 \\
Sindhi (sd)          & 14,616 & 18,556  & 33,990 \\
Sinhala (si)         & 20,363 & 29,794  & 90,866 \\
Slovak (sk)          & 232,109 & 301,681  & 4,014,344 \\
Slovenian (sl)       & 166,997 & 238,706  & 3,754,135 \\
Somali (so)          & 6,716 & 9,595  & 53,132 \\
Spanish (es)         & 1,547,372 & 3,313,727  & 37,749,593 \\
Sundanese (su)       & 54,921 & 61,716  & 598,878 \\
Swahili (sw)         & 53,926 & 74,634  & 693,049 \\
Swati (ss)           & 514 & 610  & 4,344 \\
Swedish (sv)         & 3,755,203 & 6,143,945  & 39,409,278 \\
Tagalog (tl)         & 79,036 & 181,951  & 562,526 \\
Tamil (ta)           & 129,591 & 168,718  & 1,110,037 \\
Telugu (te)          & 71,819 & 98,189  & 841,549 \\
Thai (th)            & 139,522 & 299,433  & 2,190,249 \\
Tigrinya (ti)        & 307 & 390  & 696 \\
Tswana (tn)          & 827 & 894  & 4,896 \\
Turkish (tr)         & 338,865 & 593,365  & 5,657,757 \\
Ukrainian (uk)       & 939,234 & 1,468,963  & 16,360,016 \\
Urdu (ur)            & 156,300 & 353,391  & 1,142,953 \\
Uzbek (uz)           & 132,666 & 450,865  & 764,566 \\
Vietnamese (vi)      & 1,240,324 & 1,466,573  & 10,015,209 \\
Welsh (cy)           & 106,556 & 154,043  & 1,254,901 \\
Wolof (wo)           & 1,503 & 1,969  & 7,257 \\
Xhosa (xh)           & 1,370 & 1,610  & 14,163 \\
Yoruba (yo)          & 32,304 & 42,022  & 88,032 \\
Others               & 12,613,082 & 12,613,082 & - \\
\midrule
\textbf{total}       & 53,849,351 & 89,270,463 & 777,210,183 \\
\bottomrule
\end{tabular}

}
\caption{Number of pages, entity names, and hyperlinks used in the 105 languages used for \mgenre. Entity names are more than the pages because we also includes redirections. Hyperlinks count is after filtering missed alignments to Wikidata and augmenting when there is no name in the source language.}
\label{tab:titles105}
\end{table*}

\clearpage
\begin{figure*}[t]
    \centering
    \begin{subfigure}[b]{0.48\textwidth}
        \includegraphics[height=0.85\textheight]{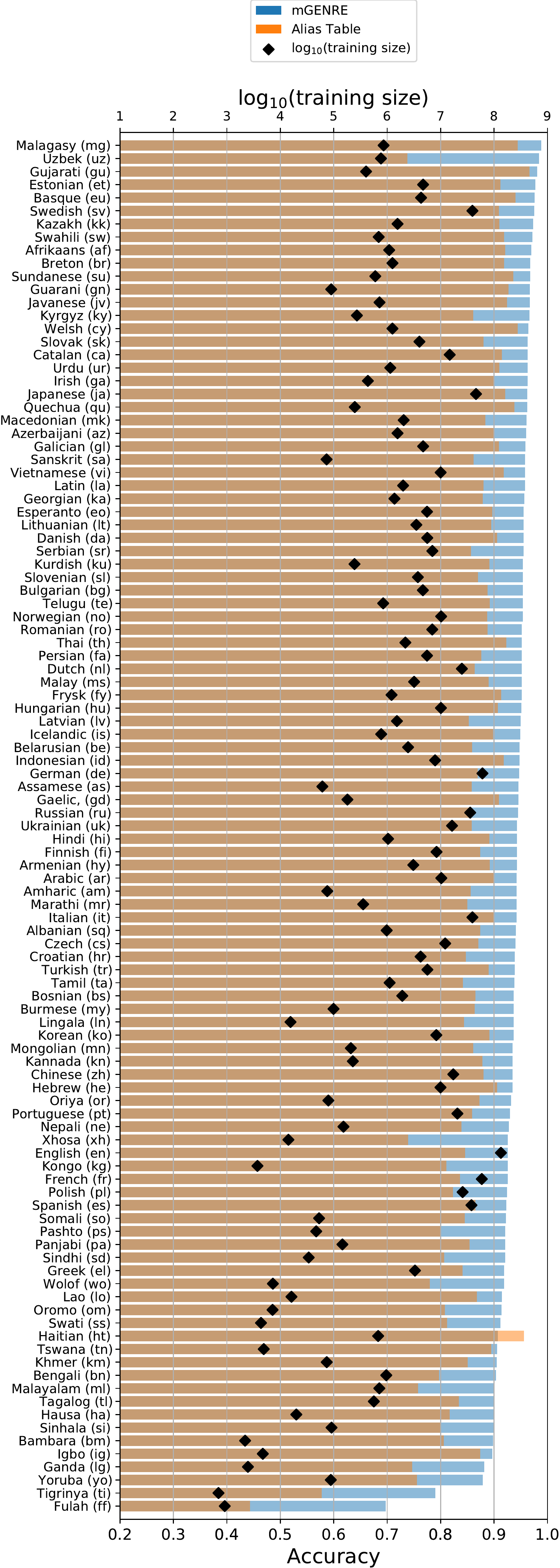}
        \caption{Sorted by \mgenre accuracy.}
    \end{subfigure}
    ~
    \begin{subfigure}[b]{0.48\textwidth}
        \includegraphics[height=0.85\textheight]{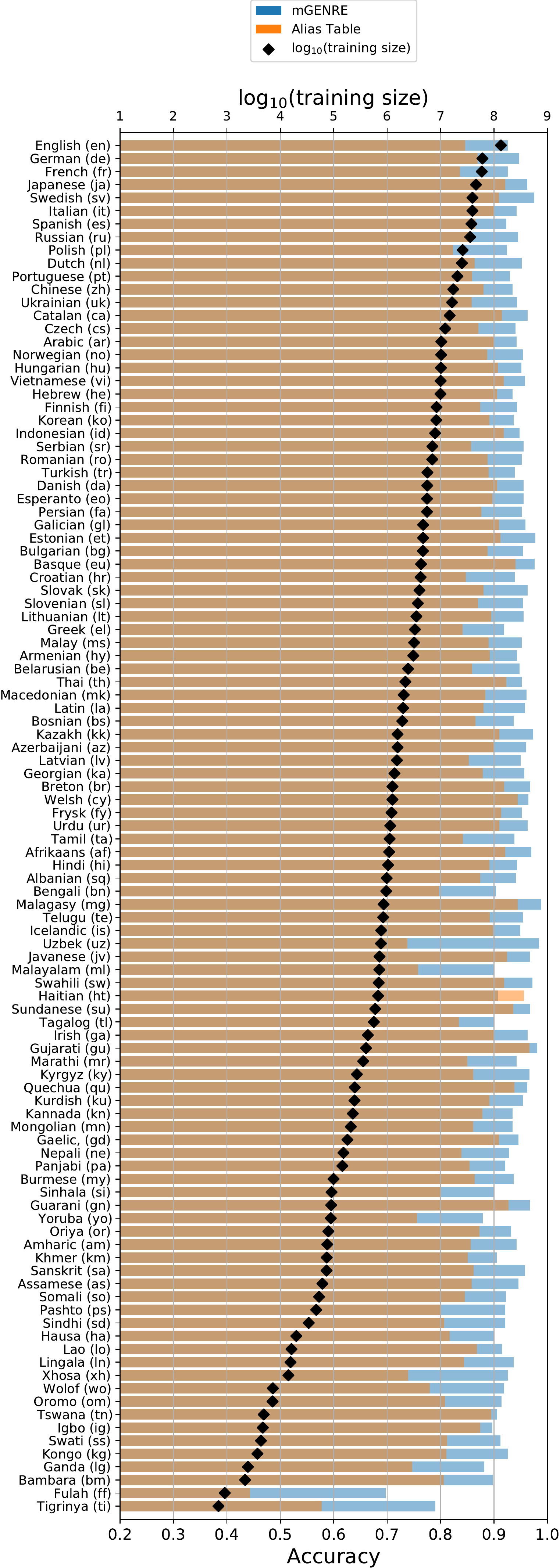}
        \caption{Sorted by training set size.}
    \end{subfigure}

    \caption{Accuracy of \mgenre and alias table on the 105 languages in our Wikipedia validation set. We report also the log-training set sizes per each language. See Table~\ref{tab:stats105} for all precise values.}
    \label{fig:stats105_langs_large}
\end{figure*}

\begin{table*}[t]
\centering
\resizebox*{!}{0.75\textheight}{  

\fontsize{8.4}{10.1}\selectfont
\begin{tabular}{lccr}
\toprule
\textbf{Language} &  \textbf{Alias Table} &  \textbf{\mgenre} &      \textbf{Support} \\
\midrule
Afrikaans (af)   &           92.1 &    97.0 &          1,089,581 \\
Albanian (sq)    &           87.4 &    94.1 &            978,394 \\
Amharic (am)     &           85.6 &    94.2 &             75,575 \\
Arabic (ar)      &           89.9 &    94.2 &         10,308,074 \\
Armenian (hy)    &           89.2 &    94.3 &          3,082,000 \\
Assamese (as)    &           85.8 &    94.6 &             61,209 \\
Azerbaijani (az) &           90.0 &    96.0 &          1,562,968 \\
Bambara (bm)     &           80.6 &    89.8 &              2,191 \\
Basque (eu)      &           94.0 &    97.6 &          4,305,648 \\
Belarusian (be)  &           85.9 &    94.8 &          2,459,794 \\
Bengali (bn)     &           79.7 &    90.4 &            960,484 \\
Bosnian (bs)     &           86.5 &    93.7 &          1,916,515 \\
Breton (br)      &           91.9 &    96.8 &          1,255,295 \\
Bulgarian (bg)   &           88.8 &    95.4 &          4,655,641 \\
Burmese (my)     &           86.4 &    93.7 &             98,992 \\
Catalan (ca)     &           91.5 &    96.3 &         14,790,419 \\
Chinese (zh)     &           88.0 &    93.5 &         17,262,417 \\
Croatian (hr)    &           84.7 &    93.9 &          4,223,179 \\
Czech (cs)       &           87.1 &    94.0 &         12,173,376 \\
Danish (da)      &           90.6 &    95.5 &          5,621,483 \\
Dutch (nl)       &           86.4 &    95.2 &         25,002,389 \\
English (en)     &           84.6 &    92.6 &        134,477,329 \\
Esperanto (eo)   &           89.7 &    95.5 &          5,570,306 \\
Estonian (et)    &           91.2 &    97.7 &          4,700,888 \\
Finnish (fi)     &           87.4 &    94.3 &          8,390,037 \\
French (fr)      &           83.6 &    92.6 &         59,006,932 \\
Frysk (fy)       &           91.3 &    95.2 &          1,206,432 \\
Fulah (ff)       &           44.3 &    69.7 &                912 \\
Gaelic, (gd)     &           90.9 &    94.6 &            180,186 \\
Galician (gl)    &           90.9 &    95.9 &          4,709,070 \\
Ganda (lg)       &           74.7 &    88.2 &              2,476 \\
Georgian (ka)    &           87.9 &    95.7 &          1,369,094 \\
German (de)      &           86.9 &    94.7 &         60,638,345 \\
Greek (el)       &           84.1 &    91.9 &          3,310,875 \\
Guarani (gn)     &           92.7 &    96.7 &             89,593 \\
Gujarati (gu)    &           96.6 &    98.1 &            402,483 \\
Haitian (ht)     &           95.6 &    90.7 &            677,064 \\
Hausa (ha)       &           81.7 &    89.9 &             19,929 \\
Hebrew (he)      &           90.6 &    93.5 &          9,947,354 \\
Hindi (hi)       &           89.1 &    94.3 &          1,040,288 \\
Hungarian (hu)   &           90.7 &    95.1 &         10,138,904 \\
Icelandic (is)   &           89.8 &    94.9 &            772,213 \\
Igbo (ig)        &           87.4 &    89.7 &              4,702 \\
Indonesian (id)  &           91.8 &    94.8 &          7,882,254 \\
Irish (ga)       &           90.1 &    96.3 &            435,135 \\
Italian (it)     &           89.9 &    94.2 &         39,382,886 \\
Japanese (ja)    &           92.1 &    96.2 &         45,957,053 \\
Javanese (jv)    &           92.4 &    96.7 &            718,589 \\
Kannada (kn)     &           87.8 &    93.5 &            227,731 \\
Kazakh (kk)      &           91.0 &    97.3 &          1,564,344 \\
Khmer (km)       &           85.1 &    90.5 &             73,950 \\
Kongo (kg)       &           81.1 &    92.6 &              3,733 \\
Korean (ko)      &           89.1 &    93.7 &          8,309,492 \\
Kurdish (ku)     &           89.1 &    95.4 &            244,779 \\
\bottomrule
\end{tabular}
}
~
\resizebox*{!}{0.75\textheight}{  

\fontsize{8.4}{10.1}\selectfont
\begin{tabular}{lccr}
\toprule
\textbf{Language} &  \textbf{Alias Table} &  \textbf{\mgenre} &      \textbf{Support} \\
\midrule
Kyrgyz (ky)      &           86.1 &    96.6 &            271,335 \\
Lao (lo)         &           86.8 &    91.5 &             16,173 \\
Latin (la)       &           88.0 &    95.8 &          1,986,307 \\
Latvian (lv)     &           85.3 &    95.0 &          1,522,814 \\
Lingala (ln)     &           84.4 &    93.7 &             15,518 \\
Lithuanian (lt)  &           89.5 &    95.5 &          3,512,764 \\
Macedonian (mk)  &           88.4 &    96.1 &          2,035,348 \\
Malagasy (mg)    &           94.4 &    98.8 &            857,000 \\
Malay (ms)       &           89.0 &    95.2 &          3,190,700 \\
Malayalam (ml)   &           75.8 &    90.0 &            712,869 \\
Marathi (mr)     &           85.0 &    94.2 &            355,536 \\
Mongolian (mn)   &           86.1 &    93.5 &            208,847 \\
Nepali (ne)      &           83.9 &    92.8 &            151,958 \\
Norwegian (no)   &           88.7 &    95.4 &         10,234,086 \\
Oriya (or)       &           87.3 &    93.2 &             79,261 \\
Oromo (om)       &           80.8 &    91.4 &              7,153 \\
Panjabi (pa)     &           85.4 &    92.1 &            145,204 \\
Pashto (ps)      &           80.0 &    92.1 &             46,987 \\
Persian (fa)     &           87.6 &    95.2 &          5,567,774 \\
Polish (pl)      &           82.3 &    92.4 &         25,817,929 \\
Portuguese (pt)  &           85.9 &    93.0 &         20,625,904 \\
Quechua (qu)     &           93.8 &    96.2 &            247,508 \\
Romanian (ro)    &           88.8 &    95.2 &          6,974,837 \\
Russian (ru)     &           85.7 &    94.5 &         35,783,391 \\
Sanskrit (sa)    &           86.2 &    95.8 &             73,380 \\
Serbian (sr)     &           85.7 &    95.5 &          7,012,202 \\
Sindhi (sd)      &           80.7 &    92.1 &             33,990 \\
Sinhala (si)     &           80.1 &    89.9 &             90,866 \\
Slovak (sk)      &           88.0 &    96.3 &          4,014,344 \\
Slovenian (sl)   &           87.0 &    95.4 &          3,754,135 \\
Somali (so)      &           84.5 &    92.2 &             53,132 \\
Spanish (es)     &           86.5 &    92.3 &         37,749,593 \\
Sundanese (su)   &           93.6 &    96.8 &            598,878 \\
Swahili (sw)     &           91.9 &    97.2 &            693,049 \\
Swati (ss)       &           81.2 &    91.2 &              4,344 \\
Swedish (sv)     &           90.9 &    97.5 &         39,409,278 \\
Tagalog (tl)     &           83.4 &    89.9 &            562,526 \\
Tamil (ta)       &           84.2 &    93.8 &          1,110,037 \\
Telugu (te)      &           89.2 &    95.4 &            841,549 \\
Thai (th)        &           92.3 &    95.2 &          2,190,249 \\
Tigrinya (ti)    &           57.8 &    79.0 &                696 \\
Tswana (tn)      &           89.5 &    90.6 &              4,896 \\
Turkish (tr)     &           89.0 &    93.9 &          5,657,757 \\
Ukrainian (uk)   &           85.8 &    94.3 &         16,360,016 \\
Urdu (ur)        &           91.0 &    96.3 &          1,142,953 \\
Uzbek (uz)       &           73.8 &    98.4 &            764,566 \\
Vietnamese (vi)  &           91.8 &    95.8 &         10,015,209 \\
Welsh (cy)       &           94.4 &    96.4 &          1,254,901 \\
Wolof (wo)       &           78.0 &    91.9 &              7,257 \\
Xhosa (xh)       &           73.9 &    92.6 &             14,163 \\
Yoruba (yo)      &           75.6 &    87.9 &             88,032 \\
\midrule
\textbf{micro-avg}        &           86.5 &    93.8 &                  - \\
\textbf{macro-avg}        &           86.6 &    93.9 &                  - \\
\textbf{total}            &           -    &    -    &        777,210,183 \\
\bottomrule
\end{tabular}

}
\caption{Accuracy of \mgenre and alias table on the 105 languages in our Wikipedia validation set. The support indicates how many datapoints where used to train where validation is done on 1,000 examples per language (less for Tigrinya and Fulah since we have less than a thousand hyperlinks).}
\label{tab:stats105}
\end{table*}

\clearpage
\begin{figure*}[t]
    \centering
    \includegraphics[width=\textwidth]{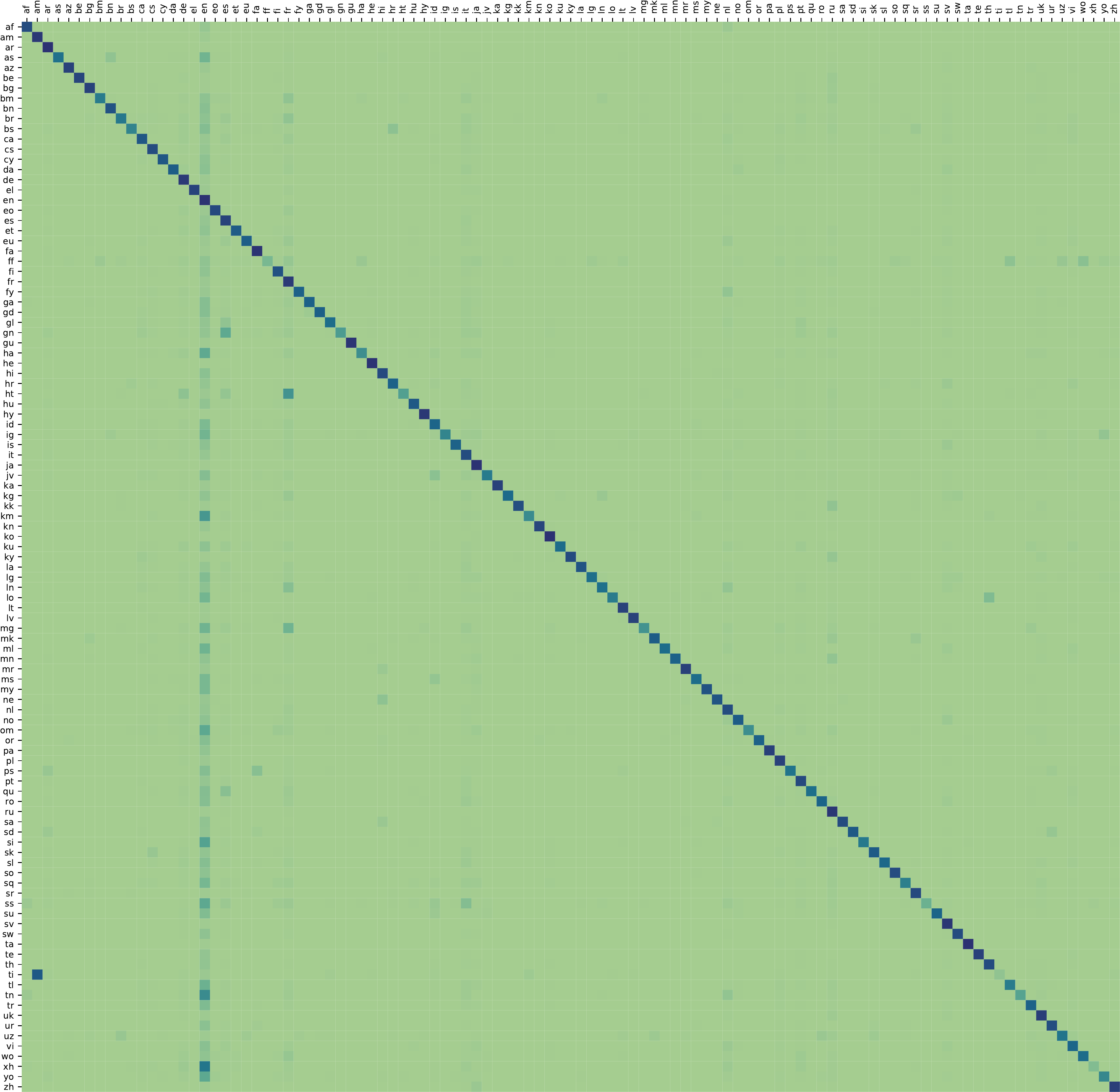}
    \caption{Distribution of languages on the top-1 prediction of \mgenre on Wikipedia heldout set. Y-axis indicates the source language where X-axis indicates the language of the top-1 prediction. The model is trained on all those languages. Clearly the model is biased to predict in the source language---note that we train in such a way---but there are some languages that are also used quite often (\eg, English). French and Russian are also often used from other languages.}
    \label{fig:heatmap_all}
\end{figure*}

\end{document}